\documentclass[10pt,journal,compsoc]{IEEEtran}

\ifCLASSOPTIONcompsoc
  \usepackage[nocompress]{cite}
\else
  \usepackage{cite}
\fi

\usepackage{graphicx}
\usepackage{subfigure}
\usepackage{booktabs} 

\usepackage{amsmath}
\usepackage{amssymb}
\usepackage{mathtools}
\usepackage{amsthm}
\usepackage{multirow} 

\usepackage{algorithmic}
\usepackage{algorithm}
\usepackage{framed}

\usepackage{url}
\usepackage{bm}
\usepackage{xcolor}
\usepackage{hyperref}
\hypersetup{
    colorlinks=true,
    linkcolor=blue,
    filecolor=magenta,      
    urlcolor=cyan,
    citecolor=green,
}

\def \xx {\bm{x}}
\def \zz {\bm{z}}

\def \hz {\bm{\widetilde{z}}}
\def \uu {\bm{\mu}}
\def \hu {\bm{\widetilde{\mu}}}
\def \oo {\bm{\omega}}

\def \SS {\bm{\Sigma}}

\def \ss {\bm{\sigma}}

\def \wu {\widetilde{\mu}}
\def \ws {\widetilde{\sigma}}

\def \pr {\mathrm{Pr}}

\newcommand{\acc}[2]{#1\scriptsize\textcolor[RGB]{168,8,13}{+#2}}
\newcommand{\macc}[2]{#1\scriptsize\textcolor[RGB]{61,145,64}{-#2}}

\newtheorem{theorem}{Theorem}[section]
\newtheorem{proposition}[theorem]{Proposition}

\begin{document}

\title{From Channel Bias to Feature Redundancy: Uncovering the “{Less is More}” Principle in Few-Shot Learning}

\author{Ji Zhang,
        Xu~Luo,
        Lianli~Gao,
        Difan~Zou,
        Heng Tao~Shen
        and~Jingkuan~Song$^{\dagger}$
\IEEEcompsocitemizethanks{
\IEEEcompsocthanksitem Ji Zhang is with the Southwest Jiaotong University, Chengdu, China. \protect\\
E-mail: jizhang.jim@gmail.com
\IEEEcompsocthanksitem Xu Luo and Lianli Gao are with the University of Electronic Science and Technology of China (UESTC), Chengdu, China.\protect\\
E-mail: frank.luox@outlook.com, lianli.gao@uestc.edu.cn
\IEEEcompsocthanksitem Difab Zou is with The University of Hong Kong, Hong Kong SAR.\protect\\
E-mail: dzou@cs.hku.hk
\IEEEcompsocthanksitem Heng Tao Shen and Jingkuan Song are with the Tongji University, Shanghai, China.
E-mail: shenhengtao@hotmail.com, jingkuan.song@gmail.com
\IEEEcompsocthanksitem Jingkuan Song is the corresponding author.
\IEEEcompsocthanksitem A preliminary version of this work has been published in ICML'22 \cite{luo2022channel}.}
}

\markboth{Journal of \LaTeX\ Class Files,~Vol.~XX, No.~X, June~2024}%
{Luo \MakeLowercase{\textit{et al.}}: Rethinking Representation Transfer in Few-Shot Learning}

\IEEEtitleabstractindextext{%
\begin{abstract}
Deep neural networks often fail to adapt representations to novel tasks under distribution shifts,
especially when only a few examples are available. This paper identifies a core obstacle behind this failure: \textit{channel bias}, where networks develop a rigid emphasis on feature dimensions that were discriminative for the source task, but this emphasis is misaligned and fails to adapt to the distinct needs of a novel task. This bias leads to a striking and detrimental consequence: \textit{feature redundancy}. We demonstrate that for few-shot tasks, classification accuracy is significantly improved by using as few as \textbf{1-5}\% of the most discriminative feature dimensions, revealing that the vast majority are actively harmful. Our theoretical analysis confirms that this redundancy originates from confounding feature dimensions—those with high intra-class variance but low inter-class separability—which are especially problematic in low-data regimes. This “\textit{less is more}” phenomenon is a defining characteristic of the few-shot setting, diminishing as more samples become available. To address this, we propose a simple yet effective soft-masking method, Augmented Feature Importance Adjustment (AFIA), which estimates feature importance from augmented data to mitigate the issue. By establishing the cohesive link from channel bias to its consequence of extreme feature redundancy, this work provides a foundational principle for few-shot representation transfer and a practical method for developing more robust few-shot learning algorithms.

\end{abstract}

\begin{IEEEkeywords}
Few-Shot Learning, Representation Learning, Transfer Learning.
\end{IEEEkeywords}}

\maketitle
\IEEEdisplaynontitleabstractindextext
\IEEEpeerreviewmaketitle

\ifCLASSOPTIONcompsoc
\IEEEraisesectionheading{\section{Introduction}\label{sec:introduction}}
\else
\section{Introduction}
\label{sec:introduction}
\fi

Deep convolutional neural networks\cite{Alexnet,Resnet} have revolutionized computer vision in the last decade, making it possible to automatically learn representations from a large number of images. The learned representations generalize effectively to novel images, enabling classification performance that is comparable to human accuracy across the majority of benchmarks. However, in addition to recognizing previously-seen categories, humans can quickly change their focus of image patterns in changing environments and recognize new categories given only a few observations. This fast learning capability, known as Few-Shot Learning (FSL), challenges current vision models on the ability to quickly adapt to novel classification tasks that are different from those in training. 

\begin{figure}[!t]
\centering
\setlength{\abovecaptionskip}{0.1cm} 
\includegraphics[width=\columnwidth]{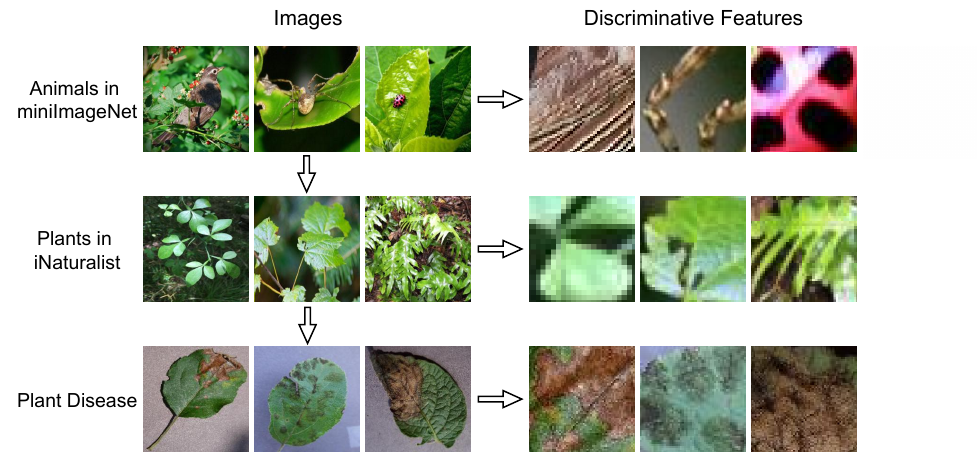}
\caption{\textbf{Illustration of Task Distribution Shift}. This figure demonstrates how the essential “discriminative features” for a FSL task change when the task distribution shifts. \textit{Top (Source Task)}: A model trained to recognize animals on the miniImageNet dataset learns to focus on features relevant to animals. \textit{Middle (Category Shift)}: When the task shifts to identifying plants in the iNaturalist dataset, the model must now focus on features of plants, which were merely background details in the original task.  \textit{Bottom (Granularity Shift)}: If the task is further refined to identifying plant diseases, the model must adapt again. It needs to learn to focus on even more fine-grained information, such as the specific patterns of lesions on a leaf, rather than the general features of the plant itself. }
\label{fig:task_shift_example}
\end{figure}

Recent studies of FSL have highlighted the importance of the quality of learned image representations~\cite{mamlrepresentation,crosstransformer,baseline}, and also showed that representations learned by neural networks do not generalize well to novel few-shot classification tasks when there is task distribution shift~\cite{metabaseline,crosstransformer,sensitivity}, where categories, domains of images, or granularity of categories in new tasks deviate from those in the training tasks. As shown in Figure \ref{fig:task_shift_example}, task distribution shift may lead to changes in discriminative image features that are critical to the classification task at hand. For example, in the source task of recognizing animals, a convolutional neural network trained on \emph{mini}ImageNet can successfully identify the discriminative information related to animals. Although the representations learned by the network may encode some plant information (from image background), plants do not appear as a main category in miniImageNet and it may be insufficient for the network to distinguish various plants in a novel few-shot task sampled from the iNaturalist dataset. Even when the network is well trained to recognize plants on iNaturalist, it is difficult to be adapted to the novel task of identifying plant diseases due to the granularity shift, since the discriminative information now becomes the more fine-grained lesion part of leaves. This fundamental mismatch between the features emphasized by the model and the features required by the new task represents a core difficulty in FSL, necessitating a closer examination of how task distribution shifts affect pre-trained representations. 


In this paper, we show for the first time that this mismatch encountered in FSL leads to a \emph{channel bias} problem in learned image representations. Specifically, in the layer after global pooling, different feature channels in the learned representation seek for different patterns~(as verified in~\cite{emergingobject,dissection}) during training, and the features are weighted (in a biased way) based on their importance to the training task.  However, when applied to novel FSL tasks, the learned image features usually do not change much or have inappropriately changed without adapting to novel tasks. This bias towards training tasks may result in imprecise attention to image features in novel tasks. To further understand this problem, we derive an oracle adjustment on the Mean Magnitude
of Channels (MMC) of image representations in binary classification tasks. Through analysis, we demonstrate that the channel bias problem exists in many different target tasks with various types of task distributions shift, and it becomes severe with the distribution shift expanding (as shown in Figure~\ref{visiualize_optimal}). Such findings naturally lead to the following profound questions: 


\begin{framed}
\textcolor{black}{
\textit{How much information in the features extracted by the pre-trained model is useful for a given downstream FSL task? Does the conclusion change with the dataset size?}}
\end{framed}

Though extensive empirical studies, we uncover a surprising phenomenon we term \textit{feature redundancy}. This is the concept that, in few-shot scenarios, the vast majority of feature dimensions are not merely unhelpful but are actively detrimental to performance: \textbf{1)} Using only the top 1-5\% most important feature dimensions often yields classification accuracy comparable to, or even better than, using the entire feature set. \textbf{2)} More strikingly, as we start with the most important features and progressively add more, performance initially improves, but after reaching an optimal point, it begins to decline. This demonstrates that most features act as “noise” or confounders when transferred to a new, low-data task, directly revealing a “\textit{less is more}” principle at the heart of FSL. By counting the frequencies of feature dimensions appearing among the top task-specific features from a pool of tasks, we further observe that different tasks usually require different task-specific feature dimensions, even when these tasks are sampled from the same dataset. These phenomena together indicate that the knowledge needed for downstream tasks can vary a lot from task to task, and there usually exists a notable mismatch between knowledge that resides in the pre-trained features and the knowledge needed for downstream tasks, at least under few-shot settings, suggesting the importance of feature selection.
Furthermore, by analyzing a toy example where features are sampled from a Gaussian distribution, we theoretically show that this extreme redundancy stems directly from feature dimensions with high intra-class variance but low inter-class separability. Such dimensions, while often ignored by classifiers trained on large datasets, act as powerful confounders in the low-data regime. This confounding effect is a defining characteristic of the few-shot setting, diminishing only as more samples become available.

To address the feature redundancy problem, we propose a simple yet effective soft-masking method named Augmented Feature Importance Adjustment (AFIA). The core insight of AFIA is to obtain a more stable and robust calculation of each feature's true importance by augmenting support images and use the augmented set to reduce estimation variance. Rather than completely removing features (hard masking), the method uses these improved importance scores to perform a ``soft adjustment'' by scaling the feature dimensions. This effectively down-weights the influence of confounding features that the model is less confident about. In doing so, AFIA provides a practical way to implement the ``less is more'' principle: it mitigates the impact of the vast, harmful majority of features, leading to consistent and significant improvements in transfer performance.

The main contributions of this work are threefold:
\begin{itemize}
\item We pioneer the identification of channel bias of pre-trained visual representations as a primary obstacle to successful few-shot transfer learning. Building on this, we uncover a significant issue of feature redundancy, where the vast majority of feature dimensions are harmful for few-shot transfer, leading to a “less is more” principle.

\item We provide rigorous theoretical and empirical analyses for both channel bias and feature redundancy problems. For the channel bias problem, we theoretically derive an oracle feature importance to help us empirically diagnose how severe the bias is. For the feature redundancy problem, our theoretical and empirical analyses together reveal its origins, its correlation with the volume of training data, and its effects on different classification methods when few samples are available.



\item Based on the analysis, we propose AFIA, a simple yet effective soft-masking method that leverages data augmentation to reduce feature redundancy effectively. Its implementation is straightforward, yet it yields consistent improvements in few-shot transfer capabilities across a wide range of pre-trained vision models and downstream datasets.


\end{itemize}

This paper is a substantial extension of our preliminary work \cite{luo2022channel}, which first identified the channel bias problem as a key obstacle in FSL. While the initial work established the existence of this bias, this paper deepens the analysis and provides a more complete narrative by making the following significant advances:

\begin{itemize}
\item \textbf{From Channel Bias to Feature Redundancy.} We move beyond identifying channel bias to uncover its more dramatic and detrimental consequence: extreme feature redundancy. We provide rigorous theoretical and empirical evidence showing that in few-shot settings, the vast majority of feature dimensions are actively harmful. This establishes the “less is more” principle as a defining characteristic of FSL

\item \textbf{A Practical and Effective Solution.} To overcome the severe feature redundancy problem, this work introduces the Augmented Feature Importance Adjustment (AFIA), a novel soft-masking method. Unlike the preliminary work, which focused only on problem diagnosis, AFIA provides a practical solution that leverages data augmentation to robustly estimate feature importance and down-weight confounding feature dimensions.

\item \textbf{Comprehensive Theoretical and Empirical Validation.} This paper provides a novel, rigorous theoretical foundation that explains the origins of feature redundancy, linking it to high-variance, confounding features that are especially problematic in low-data regimes. We supplement this theory with comprehensive experiments that verify the existence of the redundancy phenomenon and validate the consistent, significant improvements provided by AFIA across a wide range of models and benchmarks.

\end{itemize}

\section{Preliminaries}
In few-shot classification, a training set $\mathcal{D}^{train}$ is used to train a neural network parametrized by $\theta$. This model is then evaluated on a series of few-shot classification tasks constructed from a test-time dataset $\mathcal{D}^{test}$. A key assumption is the existence of a task distribution shift between $\mathcal{D}^{train}$ and $\mathcal{D}^{test}$, which can manifest as a shift in categories, domains, or task granularity \cite{guo2020broader,triantafillou2020meta}. Each evaluation task is an $N$-way $K$-shot problem. It is constructed by sampling $N$ classes from $\mathcal{D}^{test}$, and then sampling $K$ support images and $M$ query images for each class. The support set $\mathcal{S}_{\tau}=\{(x_{k,n}^{\tau},y_{k,n}^{\tau})\}_{k,n=1}^{K,N}$ is used to adapt the model or construct a task-specific classifier $p_{\theta}(\cdot|x,\mathcal{S}_{\tau})$. The classifier's performance is then measured on the query set $\mathcal{Q}_{\tau}=\{x_{m,n}^{*\tau}\}_{m,n=1}^{M,N}$. The final metric is the average prediction accuracy over many such randomly sampled tasks. 

To ensure a comprehensive evaluation across various types and degrees of task distribution shift, in the following experiments, we select a broad range of datasets for $\mathcal{D}^{train}$ and $\mathcal{D}^{test}$. For $\mathcal{D}^{train}$, we choose 1) the train split of \emph{mini}ImageNet~\cite{matchingnet} that contains 38400 images from 64 classes; 2) the train split of ImageNet 1K~\cite{ImageNet} containing more than 1M images from 1000 classes; 3) train+val split of iNaturalist 2018~\cite{iNaturalist}, a fine-grained dataset of plants and animals with a total of more than 450000 training images from 8142 classes. For $\mathcal{D}^{test}$, we choose the test split of \emph{mini}ImageNet, and all evaluation datasets of Meta-dataset~\cite{metadataset}, BSCD-FSL benchmark~\cite{BSCD} and DomainNet~\cite{DomainNet}, for a total of 19 datasets, to ensure adequate coverage of different categories, domains and task granularities.


\section{The Root of the Problem: channel bias}
\label{sec:feature_bias}

As established in the introduction, the failure of few-shot transfer often stems from a fundamental misalignment between the representation learned on a source task and the needs of a novel one. This paper posits that the root of this misalignment is a \textbf{channel bias}: the network's emphasis on different feature channels, often reflected by their average magnitude, is rigidly determined by the source task distribution. When confronted with a new task requiring a different set of discriminative features, the model's inherent and rigid feature emphasis fails to adapt, leading to a cascade of performance issues. This section will formally define this problem, provide clear empirical evidence of its existence, and analyze its core properties.

\subsection{A Motivating Observation: Universal Performance Gains from a Test-time
Simple Feature Transformation}
\label{subsec:motivating_observation}

\begin{figure}[!t]
\centering
\setlength{\abovecaptionskip}{-0cm} 
\includegraphics[width=0.6\columnwidth]{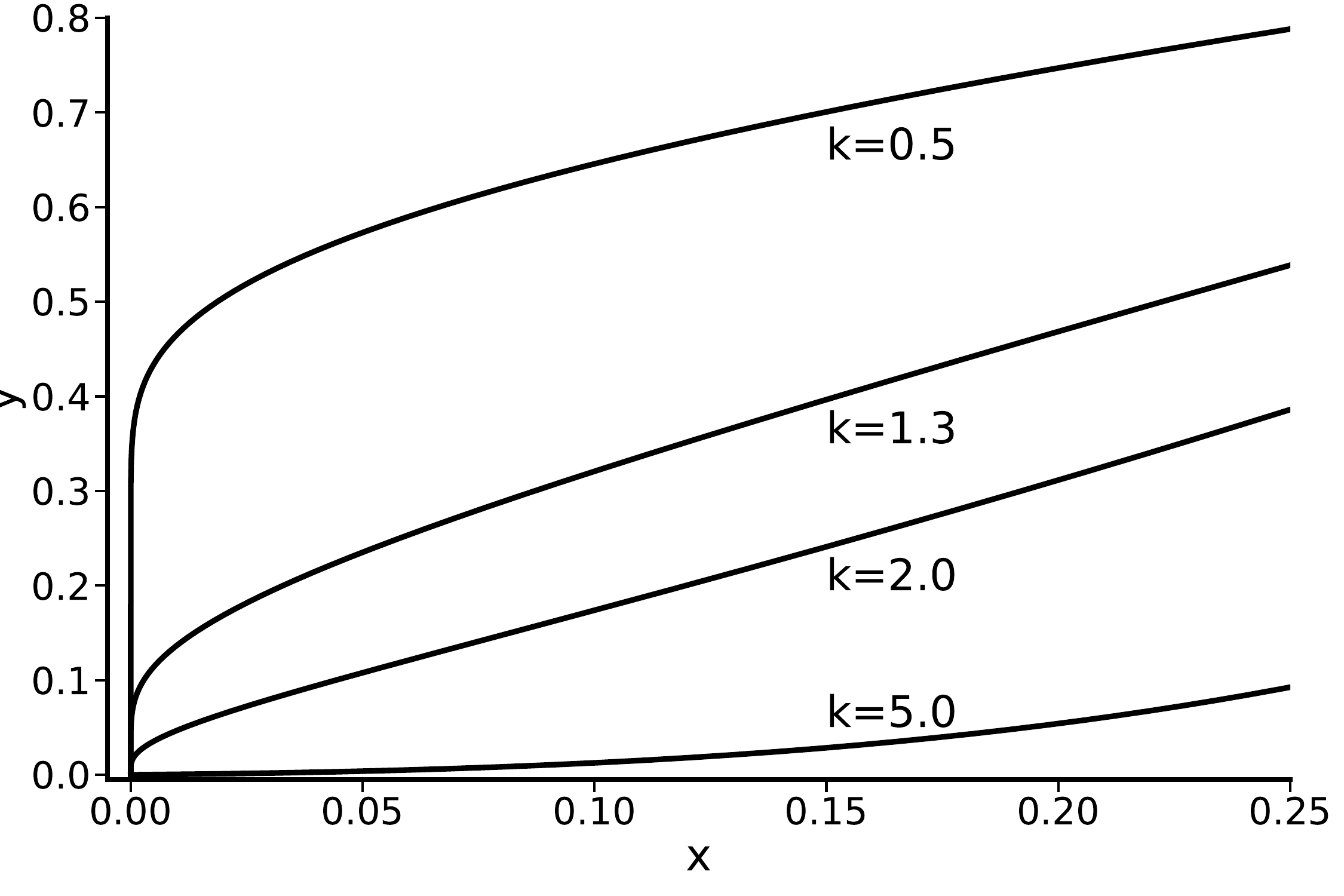}
\caption{The simple transformation function $\phi_k$ with various choices of k.}
\label{fig:transformation_plot}
\end{figure}

Let $\xx\in\mathbb{R}^D$ denote an image  and $f_\theta(\cdot)$ a feature extractor learned from the training set $\mathcal{D}^{train}$. The $l$-th channel of the feature $\zz=f_\theta(\xx)\in\mathbb{R}^d$ is defined as the $l$-th dimension of $\zz$, i.e., $\{z_i\}_{i=1}^d$ is the set of all $d$ channels. The simple transformation function $\phi_k: [0,+\infty)\rightarrow[0,+\infty)$ that we consider is defined as
\begin{equation}
    \label{simple_transformation}
    \phi_k(\lambda) =\begin{cases} \frac{1}{ln^k(\frac{1}{\lambda}+1)}, &\lambda>0\\
    0, &\lambda=0
    \end{cases}
\end{equation}
where $k>0$ is a hyperparameter. At test time, we simply use this function to transform each channel of image features, i.e.,
\begin{equation}
    \phi_k(\zz)=(\phi_k(z_1), ..., \phi_k(z_d)).
\end{equation}
When applying this transformation, we transform all image features in the target classification task regardless of whether they are in the support set or query set; any subsequent operation keeps unchanged. Note that this function can only be applied to features taking non-negative values, common in most convolutional neural networks using ReLU as the activation function. 
A plot of this function with various choices of $k$ is shown in Figure~\ref{fig:transformation_plot}. 

\begin{table*}[!t]
\setlength{\abovecaptionskip}{0.01cm}  
\caption{Performance gains of the simple feature transformation on various training and testing datasets with a broad range of choices of network architectures and algorithms. The black values indicate the original accuracy, and the red values indicate the increase. Each running of evaluation contains 10000 5-way 5-shot tasks sampled using a fixed seed, and the average accuracy is reported. The three groups of test-time datasets come from MetaDataset, BSCD-FSL benchmark and DomainNet, respectively.}
\label{tab:simple_transform_performance}
\centering
\setlength\tabcolsep{3.5pt}
\footnotesize
\begin{tabular}{c|ccccccc|ccc|c|c}
\toprule
TrainData &  \multicolumn{7}{c|}{\emph{mini}-train} & \multicolumn{3}{c|}{ImageNet} & iNaturalist & \\
Algorithm & PN & PN & CE & MetaB & MetaOpt & CE & S2M2 & PN & CE & MoCo-v2 & CE & Average\\
Architecture & Conv-4 & Res-12 & Res-12 & Res-12 & Res-12 & SE-Res50 & WRN & Res-50 & Res-50 & Res-50 & Res-50 & \\
\midrule
\emph{mini}-test & \acc{66.6}{1.2} & \acc{73.5}{2.2} & \acc{75.9}{1.6} & \acc{74.7}{2.6} & \acc{74.8}{0.5} & \acc{76.2}{0.2} & \acc{82.5}{1.2} & \macc{82.2}{1.6} & \macc{89.1}{0.5} & \acc{93.7}{2.2} & \acc{69.9}{2.2} & \acc{78.1}{1.1}\\
CUB & \acc{52.0}{2.8} & \acc{57.0}{3.0} & \acc{59.6}{2.3} & \acc{60.1}{2.6} & \acc{60.3}{1.7} & \acc{59.9}{2.2} & \acc{68.5}{2.8} & \acc{65.3}{2.5} & \acc{78.2}{0.4} & \acc{70.0}{6.8} & \acc{94.7}{0.0} & \acc{66.0}{2.5}\\
Textures & \acc{50.9}{2.3} & \acc{57.1}{4.2} & \acc{63.1}{2.4} & \acc{61.2}{3.7} & \acc{60.2}{1.8} & \acc{63.5}{0.6} & \acc{69.3}{2.9} & \acc{61.9}{2.4} & \acc{71.6}{0.8} & \acc{82.8}{0.9} & \acc{63.2}{2.3} & \acc{64.1}{2.0}\\
Traffic Signs & \acc{64.8}{2.2} & \acc{52.6}{2.1} & \acc{65.6}{1.4} & \acc{67.3}{1.5} & \acc{67.1}{4.9} & \acc{62.2}{2.9} & \acc{69.6}{3.1} & \acc{64.0}{2.2} & \acc{67.2}{3.5} & \acc{68.4}{8.8} & \acc{60.5}{4.0} & \acc{64.4}{3.3}\\
Aircraft & \acc{32.1}{0.9} & \acc{31.3}{1.6} & \acc{34.7}{1.9} & \acc{34.7}{2.3} & \acc{35.6}{2.4} & \acc{38.2}{2.0} & \acc{40.5}{4.7} & \acc{38.4}{1.7} & \acc{46.6}{2.5} & \acc{34.5}{8.8} & \acc{42.1}{2.5} & \acc{34.0}{2.9}\\
Omniglot & \acc{61.0}{10.0} & \acc{77.6}{7.8} & \acc{86.9}{3.7}& \acc{81.6}{7.9} & \acc{78.0}{9.9} & \acc{89.9}{2.3} & \acc{85.9}{7.4} & \acc{76.4}{2.9} & \acc{88.6}{5.3} & \acc{74.5}{15.8} & \acc{83.8}{9.0} & \acc{80.4}{7.5}\\
VGG Flower & \acc{71.0}{3.1} & \acc{71.1}{5.5} & \acc{79.2}{3.8} & \acc{78.3}{4.5} & \acc{78.4}{3.1} & \acc{83.0}{1.7} & \acc{87.8}{2.5} & \acc{81.4}{2.6} & \acc{89.3}{1.7} & \acc{86.2}{6.3} & \acc{91.9}{1.1} & \acc{81.6}{3.3}\\
MSCOCO & \acc{52.0}{1.2} & \acc{58.2}{1.1} & \acc{59.0}{0.7} & \acc{58.0}{1.6} & \acc{58.4}{0.1} & \acc{57.1}{0.5} & \acc{63.5}{0.1} & \macc{61.3}{0.5} & \macc{64.3}{0.4} & \acc{71.4}{1.4} & \acc{50.4}{1.9} & \acc{59.4}{0.7}\\
\midrule
Plant Disease & \acc{66.6}{7.8} & \acc{73.3}{7.9} & \acc{80.0}{5.1} & \acc{75.6}{7.6} & \acc{78.6}{4.5} & \acc{83.1}{3.2} & \acc{86.4}{3.5} & \acc{72.5}{8.0} & \acc{84.1}{3.3} & \acc{87.1}{4.7} & \acc{85.6}{4.1} & \acc{79.4}{5.4}\\
ISIC & \acc{38.5}{1.6} & \acc{36.8}{2.9} & \acc{40.4}{1.0} & \acc{38.8}{1.7} & \acc{39.5}{2.3} & \acc{37.7}{3.9} & \acc{40.5}{5.5} & \acc{39.5}{4.0} & \acc{37.8}{3.6} & \acc{43.2}{2.8} & \acc{39.0}{4.3} & \acc{39.2}{3.1}\\
EuroSAT & \acc{63.0}{4.5} & \acc{67.3}{5.5} & \acc{75.7}{2.9} & \acc{71.9}{4.5} & \acc{72.8}{5.8} & \acc{75.7}{1.6} & \acc{81.2}{2.9} & \acc{72.5}{6.1} & \acc{78.4}{2.2} & \acc{83.5}{2.7} & \acc{73.5}{3.7} & \acc{74.1}{3.9}\\
ChestX & \acc{22.9}{0.2} & \acc{23.0}{0.5} & \acc{24.1}{0.3} & \acc{23.5}{0.5} & \acc{24.5}{0.4} & \acc{23.6}{0.2} & \acc{24.2}{0.9} & \acc{23.2}{0.3} & \acc{24.2}{0.8} & \acc{25.4}{0.9} & \acc{23.9}{0.1} & \acc{23.9}{0.5}\\
\midrule
Real & \acc{67.0}{1.8} & \acc{72.2}{3.1} & \acc{76.3}{1.6} & \acc{75.0}{2.6} & \acc{75.8}{1.1} & \acc{76.7}{0.5} & \acc{81.7}{1.9} & \acc{80.5}{0.4} & \macc{87.1}{0.1} & \acc{88.8}{2.1} & \acc{72.9}{1.7}& \acc{77.6}{1.5}\\
Sketch & \acc{42.6}{2.9} & \acc{45.3}{5.0} & \acc{51.1}{2.6} & \acc{50.2}{3.4} & \acc{50.6}{2.0} & \acc{50.9}{2.4} & \acc{56.8}{4.1} & \acc{53.1}{1.5} & \acc{63.2}{2.5} & \acc{63.9}{5.8} & \acc{51.9}{1.4} & \acc{52.7}{3.1}\\
Painting & \acc{49.0}{1.7} & \acc{52.5}{3.3} & \acc{56.1}{1.4} & \acc{55.1}{2.5} & \acc{56.2}{0.7} & \acc{59.3}{0.8} & \acc{64.2}{1.8} & \macc{61.8}{0.2} & \acc{69.6}{0.5} & \acc{76.5}{3.0} & \acc{56.4}{1.9}& \acc{59.7}{1.6}\\
Clipart & \acc{47.5}{3.6} & \acc{49.7}{4.8} & \acc{55.5}{3.1} & \acc{54.9}{4.3} & \acc{56.4}{2.6} & \acc{60.4}{2.3} & \acc{63.0}{4.3} & \acc{60.9}{1.8} & \acc{72.7}{1.5} & \acc{67.4}{7.0} & \acc{58.4}{2.2} & \acc{58.8}{3.4}\\
\bottomrule
\end{tabular}
\end{table*}

\begin{figure}[!t]
\setlength{\abovecaptionskip}{-0cm} 
\centering
\includegraphics[width=0.76\columnwidth]{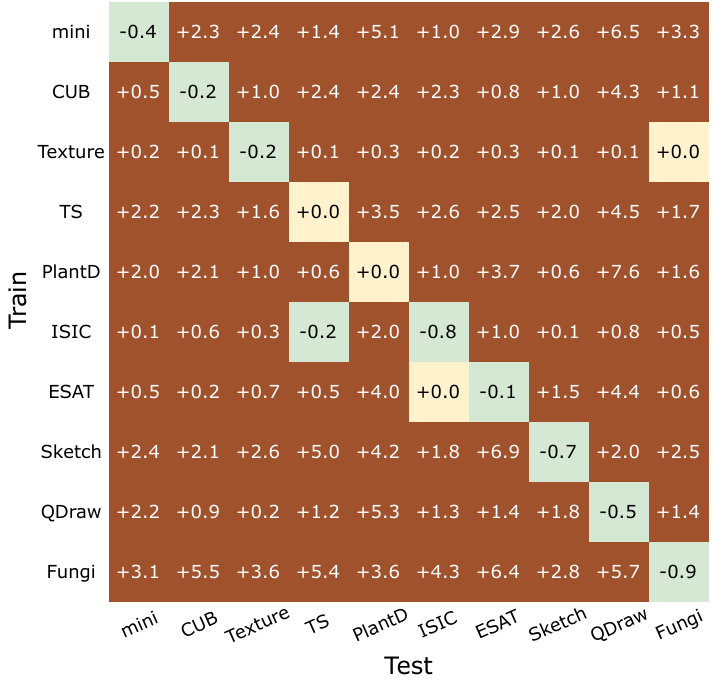}
\caption{In-distribution (diagonal) and out-of-distribution (off-diagonal) performance gains of the simple channel-wise transformation on representations trained with CE. When the test-time dataset equals the training dataset (diagonal), the categories of images remain the same but test-time images are unseen during training.}
\label{fig:pairwise}
\end{figure}

Table \ref{tab:simple_transform_performance} shows the remarkable effectiveness of this simple transformation on 5-way 5-shot tasks across a wide spectrum of setups (training and evaluation details are in \textbf{Appendix A}). We test multiple algorithms, including conventional training (CE \cite{he2016deep}, S2M2 \cite{mangla2020charting}), meta-learning (ProtoNet (PN) \cite{snell2017prototypical}, Meta-Baseline \cite{chen2021meta}, MetaOpt \cite{lee2019meta}), and self-supervised learning (MoCo-v2 \cite{he2020momentum}). We also use various backbone architectures (Conv-4, ResNet-12, ResNet-50, etc.). With a fixed hyperparameter $k=1.3$, the transformation yields substantial improvements (0.5-7.5\% on average) on nearly all out-of-distribution test sets (we show how performance varies with different $k$ values in \textbf{Appendix B}). The exceptions are informative: performance slightly degrades when the training and testing distributions are very similar (e.g., training on ImageNet and testing on \emph{mini}ImageNet or MSCOCO). This suggests the transformation is specifically beneficial for handling task distribution shift. To verify this, we conduct an in-distribution vs. out-of-distribution experiment in Figure \ref{fig:pairwise}. The results are clear: the transformation only provides a benefit when the test-time data distribution deviates from the training one (the off-diagonal entries), confirming its role in aiding representation transfer under distribution shift. This shift can be in domain (Sketch to QuickDraw), category (Plant Disease to Fungi), or granularity (iNat. to Plant Disease).



\subsection{Analysis of the Simple Transformation}
\label{sec3}
\begin{figure}[!t]
\setlength{\abovecaptionskip}{-0cm} 
\centering
\includegraphics[width=\columnwidth]{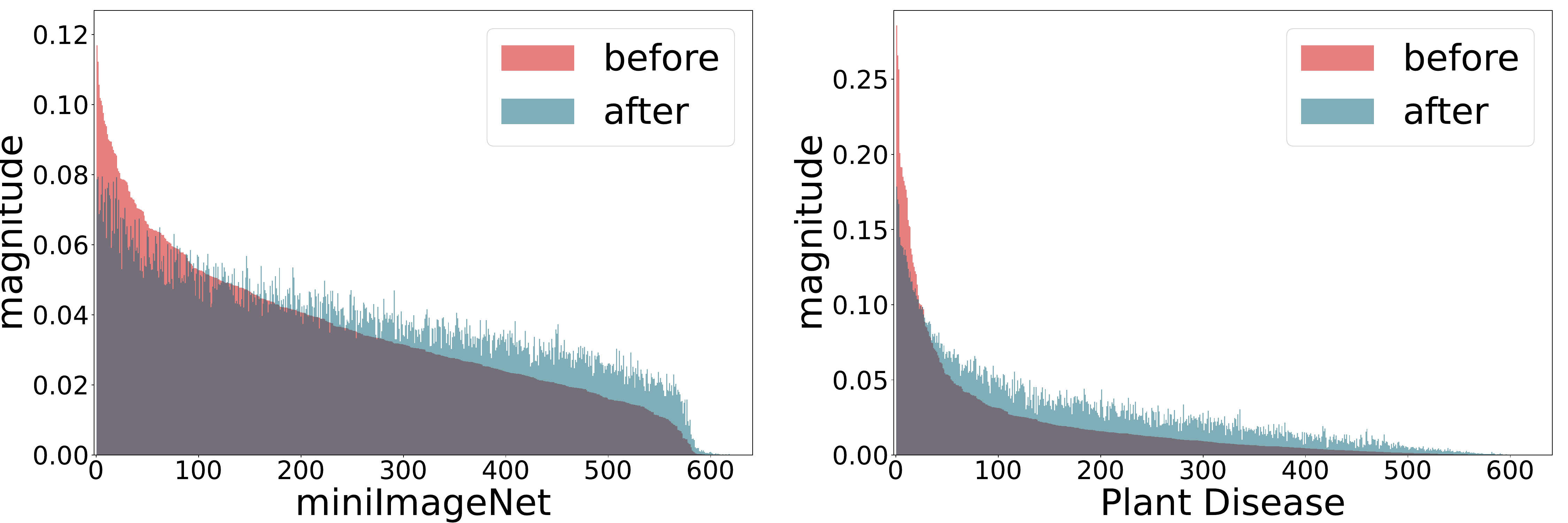}
\caption{Mean magnitudes of feature features before and after applying the simple transformation. The feature extractor is trained using PN on the training set of \emph{mini}ImageNet. Left: test set of \emph{mini}ImageNet. Right: The Plant Disease dataset. The change of relative magnitude is due to different variances of features.}
\label{fig:feature_visualization}
\end{figure}

Here, we analyze the simple transformation, which leads us to discover the channel bias problem of visual representations.
Given the transformation function described in Eq.(\ref{simple_transformation}), it can be first noticed that
\begin{gather}
\label{property}
\phi_k'(\lambda)>0, \lim_{\lambda\to 0^{+}}\phi_k'(\lambda)=+\infty,\notag\\
\exists t>0, \quad s.t. \quad \forall \lambda\in(0,t), \phi_k''(\lambda)<0,
\end{gather}
where $t$ is a large value for most $k$, relative to the magnitudes of almost all channels (e.g., when $k=1.3$, $t\approx 0.344$, while most channel values are less than $0.3$). The positiveness of the derivative ensures that the relative relationship between channels will not change,  while the negative second derivative narrows their gaps; the infinite derivative near zero pulls up small channels by a large margin, i.e., $\lim_{\lambda\to 0^+}\frac{\phi_k(\lambda)}{\lambda}=+\infty$. See \textbf{Appendix C} for the necessity of all these properties. A clear impact of these properties on features is to make channel distribution smooth: suppress channels with high magnitude, and largely amplify channels with low magnitude. 
This phenomenon is clearly shown in Figure \ref{fig:feature_visualization}, where we plot mean magnitudes of all 640 feature channels on \emph{mini}ImageNet and PlantDisease, with red ones being the original distribution, blue ones being the transformed distribution. The transformed distribution becomes more uniform.

Intuitively, different channels have high responses to different features, and a larger Mean Magnitude of a Channel (MMC) implies that the model puts more emphasis on this channel, hoping that this channel is more important for the task at hand. Combining the analysis above with previous experiment results, we conjecture that the MMC of representations should change when testing on novel tasks with a shift in distribution. This meets our intuition that different tasks are likely to be characterized by distinct discriminative features, as shown in the examples of Fig \ref{fig:task_shift_example}.

\begin{table*}[t]
\setlength{\abovecaptionskip}{0.01cm}  
\setlength\tabcolsep{4.2pt}
\footnotesize
\caption{The performance gains of the oracle MMC on 5-shot binary classification tasks on various datasets. The derived MMC improves the few-shot performance of both metric and non-metric test-time methods: Nearest-Centroid Classifier (NCC) and Linear Classifier (LC).}
\label{Optimal}
\centering
\begin{tabular}{ccc|cccccccccc|c}
\hline
 Algorithm & Classifier & Transformation & mini & CUB & Texture & TS & PlantD & ISIC & ESAT & Sketch & QDraw & Fungi & Avg
\\
\hline
\multirow{3}{*}{PN}& \multirow{3}{*}{NCC} &None & 90.5 & 80.6 & 80.6 & 85.1 & 89.2 & 65.7 & 86.5 & 71.9 & 82.4 & 74.6 & 80.7
\\
& & Simple& 91.3 & 82.4 & 83.1 & 85.8 & 93.0 & 68.6 & 89.2 & 75.2 & 85.1 & 77.2 & 83.1
\\
& &Oracle  & \textbf{93.1} & \textbf{88.7} & \textbf{87.2} & \textbf{92.4} & \textbf{95.6} & \textbf{69.1} & \textbf{91.5} & \textbf{81.2} & \textbf{89.4} & \textbf{88.4} & \textbf{87.7}
\\ \hline
\multirow{3}{*}{S2M2}&\multirow{3}{*}{LC}&None & 94.0 & 87.1 & 85.7 & 88.7 & 95.0 & 68.7 & 93.5 & 78.7 & 85.5 & 82.8 & 86.0
\\
& &Simple & 94.4 & 88.3 & 87.3 & 91.2 & 96.4 & 72.2 & 93.8 & 81.0 & 89.2 & 84.5 & 87.8
\\
& &Oracle & \textbf{96.3} & \textbf{94.0} & \textbf{90.7} & \textbf{96.1} & \textbf{98.3} & \textbf{72.6} & \textbf{95.2} & \textbf{87.0} & \textbf{93.0} & \textbf{93.3} & \textbf{91.7} \\
\hline
\end{tabular}
\end{table*}

\begin{figure}[t]
\setlength{\abovecaptionskip}{-0cm} 
\centering
\centerline{\includegraphics[width=0.98\linewidth]{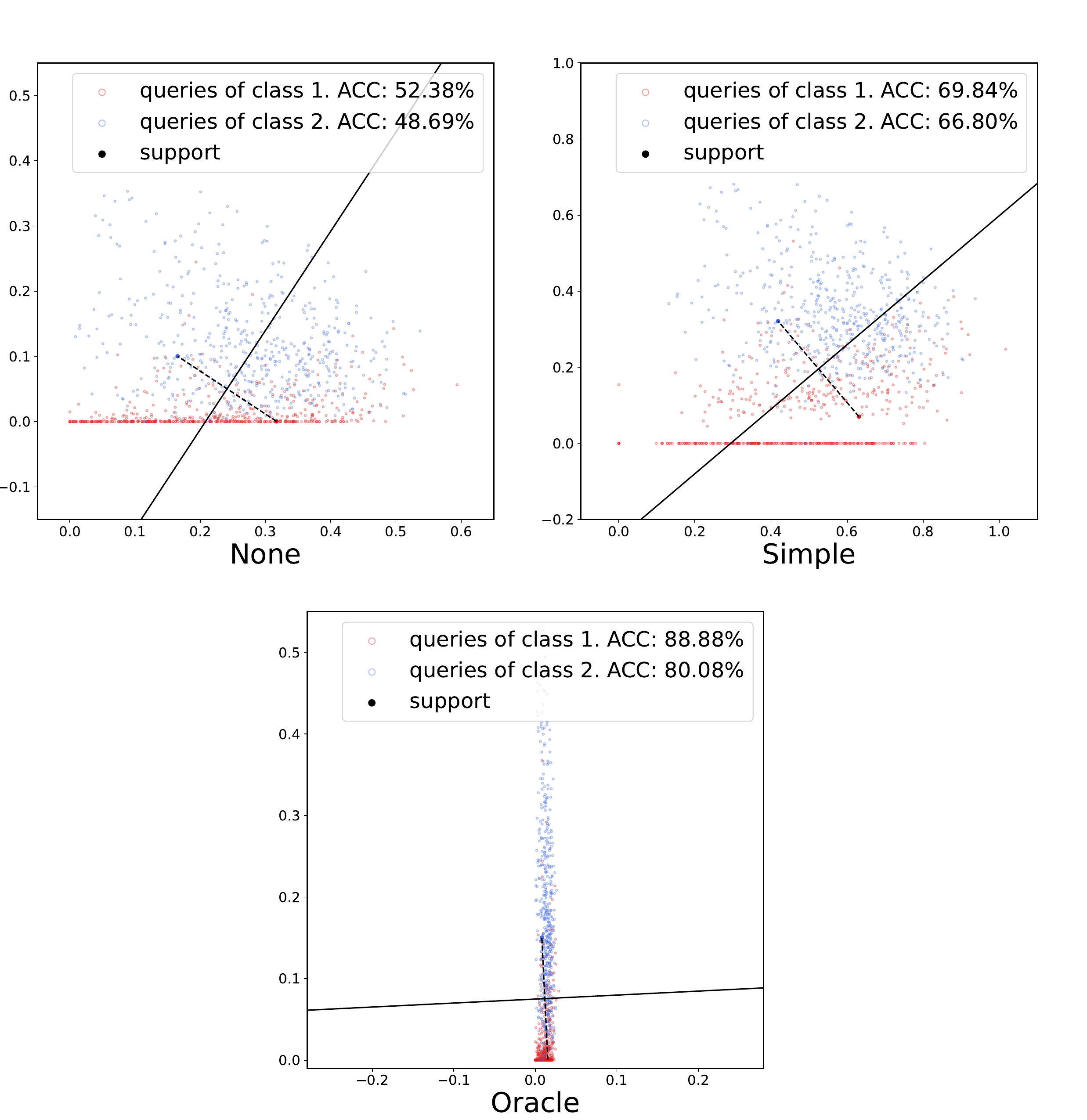}}
\caption{\textbf{Visualization of two channels of image features in two classes of Plant Disease.} The feature extractor is trained using PN on \emph{mini}ImageNet. We visualize a one-shot task with only the two channels available for classification. The plot with ``None'' shows the original channels. The plots with ``Simple'' and ``Oracle'' show channels adjusted by the simple and oracle transformation. The per-class accuracy is calculated as the proportion of samples that are correctly classified by the classification boundary in each class.}
\vskip -0.15in
\label{gaussion_example}
\end{figure}

\subsection{Deriving the Oracle MMC of Any Binary Task}

We now wonder how much the  MMC estimated by neural networks in a task deviates from the best MMC or \emph{channel importance} of that task. To achieve this goal, we first derive the optimal MMC for any classification task by multiplying a positive constant to each channel of features, given that we know the first-order and second-order statistics of features. For convenience, we consider the binary classification problem. Specifically, let $\mathcal{D}_1$, $\mathcal{D}_2$ denote probability distributions of two classes over feature space $\mathcal{Z}\subset [0,+\infty)^d$, and $\zz_1\sim\mathcal{D}_1$, $\zz_2\sim\mathcal{D}_2$ denote samples of each class. Let $\uu_1, \uu_2$ and $\SS_1, \SS_2$ denote their means and covariance matrices, respectively. We assume that the channels of features are uncorrelated with each other, i.e., there exist $\ss_1, \ss_2\in[0,+\infty)^d$, \emph{s.t.}  $\SS_1=\mathrm{diag}(\ss_1)$, $\SS_2=\mathrm{diag}(\ss_2)$. The original MMC of the binary task is defined as $\oo^o=(\uu_1+\uu_2)/2$.
We assume that the MMC after adjustment is $\oo\in [0,+\infty)^d$. Let $\hz_1$, $\hz_2$ denote standadized version of $\zz_1$, $\zz_2$ that have unit MMC, i.e., $\widetilde{z}_{1,l}=z_{1,l}/\omega^o_{l},\widetilde{z}_{2,l}=z_{2,l}/\omega^o_{l}\Rightarrow(\wu_{1,l}+\wu_{2,l})/2=1, \forall l\in [d]$ ($[d]$ is equivalent to $\{1,2,...,d\}$). A simple approach to adjust MMC to $\oo$ is to transform features to $\oo\odot \hz_1$ and $\oo\odot\hz_2$ respectively, where $\odot$ denotes the hadamard product. Here, we consider a metric-based classifier. Specifically, a standardized feature $\bm{\hz}$ is classified as the first class if $||\oo\odot (\hz-\hu_1)||_2<||\oo\odot(\hz-\hu_2)||_2$ and otherwise the second class. This classifier is actually the Nearest-Centroid Classifier (NCC)~\cite{ProtoNet} with accurate centroids. Assume that two classes of images are sampled equal times, then the expected misclassification rate of this classifier is
\begin{align}
\mathcal{R} = 
        \frac{1}{2}&[\mathbb{P}_{\zz_1\sim \mathcal{D}_1}(||\oo\odot (\hz_1-\hu_1)||_2>||\oo\odot (\hz_1-\hu_2)||_2)\nonumber\\
        +&\mathbb{P}_{\zz_2\sim \mathcal{D}_2}(||\oo\odot (\hz_2-\hu_2)||_2>||\oo\odot (\hz_2-\hu_1)||_2)].
\end{align}
The following theorem gives an upper bound of the misclassification rate and further gives the \emph{oracle} MMC of any given task.
\begin{proposition}
\label{thm:bigtheorem}
Assume that $\mu_{1,l}\neq\mu_{2,l}$ and $\sigma_{1,l}+\sigma_{2,l}>0$ hold for any $l\in[d]$, then we have
\begin{equation}
    \begin{split}
\mathcal{R}\leq\frac{8\sum_{l=1}^d\omega_l^4(\ws_{1,l}+\ws_{2,l})^2}{(\sum_{l=1}^d\omega_l^2(\wu_{1,l}-\wu_{2,l})^2)^2}.
    \end{split}
\end{equation}
To minimize this upper bound, the adjusted oracle MMC of each channel $\omega_l$ should satisfy:
\begin{equation}
\label{oracle_adjustment}
\omega_l \propto \frac{|\mu_{1,l}-\mu_{2,l}|}{\sigma_{1,l}+\sigma_{2,l}}.
\end{equation}
\end{proposition}

\begin{table*}[t]
\setlength{\abovecaptionskip}{0.01cm}  
\setlength\tabcolsep{2.0pt}
\footnotesize
\caption{Three levels of distance  between different MMCs or $l_1$-normalized image features. The first row shows the dataset-level distance between the original MMC of the training
set (mini-train) and each test set; the second row shows the dataset-level distance between the original and oracle MMCs on each dataset; rows 3-6 show the task-level and image-level distances (both amplified by $10^6$ times) between MMCs obtained by simple and oracle transformation or between original MMCs (None) and the MMCs obtained by oracle transformation. The feature extractor is trained using PN on the training set of \emph{mini}ImageNet (mini-train).}
\label{distance}
\centering
\begin{tabular}{ccc|ccccccccccc}
\\[-1em]
\hline
 Level & Compared dataset & Trans. & mini-train & mini-test & CUB & Texture & TS & PlantD & ISIC & ESAT & Sketch & QDraw & Fungi\\\hline
 \multirow{2}{*}{Dataset} & Train \emph{v.s.} Test & None & - & 0.18 & 1.56 & 0.88 & 1.13 & 1.54 & 2.28 & 1.30 & 1.01 & 1.58 & 0.79\\
 &Test & None \emph{v.s.} Oracle & 0.42 & 0.72 & 3.60 & 1.78 & 4.04 & 3.92 & 3.47 & 5.62 & 4.26 & 3.37 & 3.87\\\hline
 \multirow{2}{*}{Task}&\multirow{2}{*}{Test} & None \emph{v.s.} Oracle & 3.60 & 4.04 & 3.53 & 4.13 & 3.68 & 4.09 & 3.15 & 4.24 & 5.31 & 4.22 & 3.38\\
 & & Simple \emph{v.s.} Oracle & 3.54 & 3.80 & 2.93 & 3.62 & 3.22 & 3.65 & 2.59 & 3.71 & 4.35 & 3.18 & 2.78\\\hline
 \multirow{2}{*}{Image}&\multirow{2}{*}{Test} & None \emph{v.s.} Oracle & 10.52 & 10.65 & 11.53 & 25.20 & 9.88 & 9.75 & 13.04 & 16.33 & 27.36 & 13.46 & 11.39\\
 & & Simple \emph{v.s.} Oracle & 7.98 & 8.14 & 8.69 & 16.74 & 7.06 & 7.34 & 8.43 & 11.22 & 19.50 & 9.32 & 8.71\\\hline
\end{tabular}
\end{table*}

\begin{figure*}[t]
\setlength{\abovecaptionskip}{-0cm} 
\centering
\centerline{\includegraphics[width=\linewidth]{Images/best_avg_all_p1.pdf}}
\caption{{Visualization of MMC of ten datasets  $\oo_D$ before and after the use of simple and oracle transformation.} In each plot, a point represents a channel, and the x-axis and y-axis represent the MMC before and after transformation respectively, averaged over all possible binary tasks in the corresponding dataset. For comparison, we also plot the line $y=x$ representing the ``None'' scenario where none of the transformations are applied to features. The feature extractor is trained using PN on \emph{mini}ImageNet.}

\vskip -0.15in
\label{visiualize_optimal}
\end{figure*}

Proofs are given in \textbf{Appendix D}. We here use the word ``oracle'' because it is derived using the class statistics of the target dataset, which is not available in few-shot tasks. This derived MMC has an intuitive explanation: if the difference between the means of features from two classes is large but the variances of features from two classes are both small, the single channel can better distinguish the two classes and thus should be emphasized in the classification task. 
In fact, if we further assume $x_{1,l}$ and $x_{2,l}$ are Gaussian-distributed and consider only using the $l$-th channel for classification, then the misclassification error for the $i$-th class ($i=1,2$) is a strictly monotonically decreasing function of $|\mu_{1,l}-\mu_{2,l}|/\sigma_{i,l}$. 

Table~\ref{Optimal} shows the performance improvement over the simple feature transformation when adjusting the MMC to derived oracle one in each of the real few-shot binary classification tasks. For every sampled binary task in a dataset, we calculate the oracle adjustment based on Eq.~(\ref{oracle_adjustment}); see \textbf{Appendix E} for details. The oracle MMC improves performance on all datasets, and always by a large margin. Note that although the oracle MMC is derived using a metric-based classifier, it can also help a linear classifier to boost performance, which will be further discussed in Section 4. The large performance gains using the derived channel performance indicate that the MMC of features on new test-time few-shot task indeed has a large mismatch with ground-truth channel importance. 

To obtain a better understanding, in Figure \ref{gaussion_example}, we visualize image representations of two classes when transferred from \emph{mini}ImageNet to Plant Disease. The two exhibited classes are apples with Apple Scab and Black Rot diseases, respectively. We visualize 2 out of 640 channels in the features, shown as the $x$-axis and $y$-axis in the figure. We select these channels by first selecting a channel that requires a large suppression of MMC ($x$-axis), and then a channel that requires a large increase ($y$-axis). As seen, the $x$-axis channel has a large intra-class variance (
the variances are 0.13 and 0.11 in two classes on the $x$-axis channel, compared to 0.03 and 0.08 on the $y$-axis channel) and a small class mean difference (about 0.03, compared to 0.13 on the $y$-axis channel), so it is hard to distinguish two classes through this channel. By adjusting the mean magnitude of this channel, the simple transformation and oracle adjustment decrease the intra-class variance of the $x$-axis channel, and so decrease its influence on classification. Similarly, the $y$-axis channel can better distinguish two classes due to its relatively larger class mean difference and smaller intra-class variance, so the influence of the $y$-axis channel should be strengthened.

\subsection{Analysis of Channel Importance}
Next, we take the derived oracle MMC as an approximation of the ground-truth channel importance, and use it to observe how the simple transformation works, as well as how much the channel emphasis of neural networks deviates from the ground-truth channel importance of tasks in each test-time dataset. We define MMC of a dataset $D$ as the average $l_1$-normalized MMCs over all possible binary tasks in that dataset. Specifically, suppose in one dataset $D$ there are $C$ classes, and let $\oo_{ij}$ denote the MMC in the binary task discriminating the $i$-th and $j$-th class. $\overline{\oo_{ij}}=\oo_{ij}/||\oo_{ij}||_1$ normalizes the MMC, such that the $l$-th component of the vector $\overline{\oo_{ij}}$ represents the percentage of channel emphasis on the $l$-th channel. Then the MMC of $D$ is defined as $\oo_D=\overline{\sum_{1\leq i<j\leq C}\overline{\oo_{ij}}}$,
which gives average percentages of channel emphasis over all binary tasks. We visualize the oracle MMC, compared with MMC adjusted by the simple transformation and the original MMC of each dataset in Figure \ref{visiualize_optimal}. A point in each figure represents a channel of the image features, with $x$ and $y$ axis being its MMC of that dataset before and after transformation, respectively. To obtain a more precise understanding, we also want to quantitatively measure difference between different MMCs or image features. To achieve this, given a distance measure $d(\cdot, \cdot)$ (not necessarily a metric), we define three levels of distances: (1) dataset-level distance $d(\oo_{D_a}, \oo_{D_b})$ that measures the distance between MMCs of two datasets (or the same dataset with different transformations); (2) in-dataset task-level distance $\frac{C(C+1)}{2}\sum_{1\leq i<j \leq C}d(\overline{\oo_{ij}^{a}},\overline{\oo_{ij}^{b}})$ that measures average distance between MMCs of all tasks 
from a dataset obtained by different feature transformations, and (3) image-level distance $\frac{1}{|D|}\sum_{i=1}^{|D|}d(\overline{\zz_a^i},\overline{\zz_b^i})$, a more fine-grained one that measures average distance between all $l_1$-normalized image features $\overline{\zz_a^i},\overline{\zz_b^i}$ of dataset $D$ under different feature transformations. For dataset-level distance, we adopt the normalized mean square difference $d(\xx,\bm{y})=\frac{1}{d}\sum_{l=1}^d(x_l-y_l)^2/x_l^2$, since it treats each channel equally w.r.t. to the scale and is sensitive to high deviation. However, for task-level and image-level distance, we choose the mean square difference  $d(\xx,\bm{y})=\frac{1}{d}\sum_{l=1}^d(x_l-y_l)^2$ instead to avoid high variations caused by a single task or image feature that has channels with very small magnitude. We calculate the distance (1) between the original MMC of the training set (mini-train) and each test set, to see how much neural networks change channel emphasis when faced with novel tasks, (2) between the original and oracle MMC to see how much the changed emphasis is biased on each dataset, and (3) between the simple and oracle MMC of each dataset to see how much the simple transformation alleviates the problem. The results are shown in Table \ref{distance}. 



\textbf{Neural networks are overconfident in previously learned channel importance.} Comparing the first and second rows in Table \ref{distance}, we can see that the adjustment of MMC that the network made on new tasks is far from enough: the distance of original MMCs between train and test set (the first row) is much smaller than that between original and oracle MMCs on the test set. This suggests channels that are important to previously learned tasks are still considered by the neural network to be important for distinguishing entirely new tasks, but in fact, the discriminative channels are very likely to change on new tasks. This can be also observed from each plot in Figure \ref{visiualize_optimal}, where the oracle MMC pushes up channels having small magnitudes and suppresses channels having large magnitudes. The magnitudes of a large number of small-valued channels are amplified 10$\times$ times or more by the oracle MMC, while large-valued channels are suppressed 5$\times$ times or more, and in most datasets originally large-valued channels eventually have similar channel importance to those of originally small-valued channels. The simple transformation, although not being perfect, also regularizes channels due to its smoothing property discussed in Section 3. We call this problem the \emph{channel bias} problem.

\begin{figure}[t]
\setlength{\abovecaptionskip}{-0cm} 
    \centering
\centerline{\includegraphics[width=0.98\linewidth]{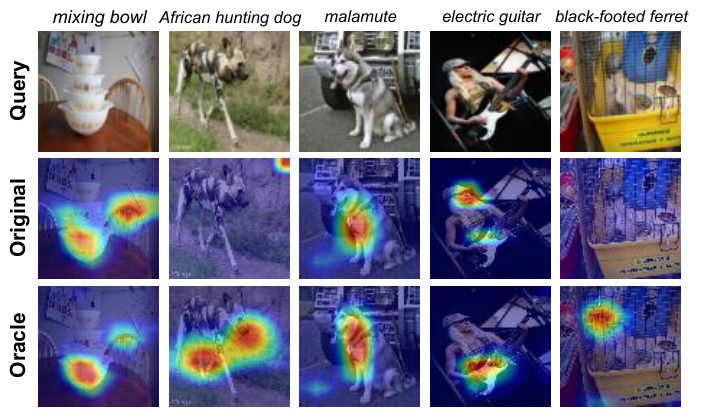}}
\caption{Examples of Grad-Cam~\cite{grad-cam} class activation maps of query samples using PN before and after the oracle adjustment of MMC on binary 5-shot tasks sampled from the test set of \emph{mini}ImageNet.}
\vskip -0.1in
\label{cam}
\end{figure}

\textbf{The channel bias problem diminishes as task distribution shift lessens.} The channel patterns in Figure \ref{visiualize_optimal} on all datasets look similar, except for \emph{mini}ImageNet, whose overall pattern is close to the line $y=x$ representing the original MMCs. There does not exist \emph{dominant} channels when testing on \emph{mini}ImageNet~(The maximum scale of channels is within 0.006), while on other datasets there are channels where the neural network assigns much higher but wrong MMCs which deviate far away from the $y = x$ line. In the second row of Table \ref{distance}, we can also see that the distance between the original and oracle MMCs on \emph{mini}ImageNet, especially on \emph{mini}-train that the model trained on, is much smaller than that on other datasets.
Since \emph{mini}-test has a similar task distribution with \emph{mini}-train, we can infer that the channel bias is less serious on datasets that have similar task distribution. This explains why in Table \ref{tab:simple_transform_performance} and Figure \ref{fig:pairwise} the simple transformation gets a relatively low improvement when trained on \emph{mini}-train and tested on \emph{mini}-test, and even degrades performance when trained and tested on tasks sampled from the same task distribution.

\textbf{The channel bias problem distracts the neural network from new objects.} In Figure \ref{cam}, we compare some class activation maps before and after the oracle adjustment of MMC. We observe that adjusting channel importance helps the model adjust the attention to the objects responsible for classification using a classifier constructed by only a few support images. This matches observation in previous work~\cite{emergingobject,dissection} that different channels of image representations are responsible for detecting different objects. The task distribution shift makes models confused about which object to focus on, and a proper adjustment of channel emphasis highlights the objects of interest.

\textbf{The simple transformation pushes MMCs towards the oracle ones.} Observing Figure \ref{visiualize_optimal}, it is evident that the simple transformation pushes MMCs towards the oracle ones (compared with the line $y=x$), albeit not perfectly. This observation is further confirmed by the None \emph{v.s.} Oracle and Simple \emph{v.s.} Oracle comparison  of fine-grained task-level and image-level distance shown from the third row to the last row of Table  \ref{distance}. On each of the test-time dataset, the distance between MMCs obtained by simple and oracle transformation is significantly smaller than that bewteen original MMCs and the MMCs obtained by  oracle transformation.






\subsection{Analysis of the Number of Shots}

We have seen that the channel bias problem is one of the main reasons why image representations cannot generalize well to new few-shot classification tasks. However, two questions remain to be answered: (1) we are still unclear whether this problem is only tied with few-shot image classification. In all previous experiments, we tested on tasks where only 5 labeled images per class are given. What will happen if we have more training examples in the new task? (2) How much will different test-time methods be influenced by the channel bias problem? If we have the opportunity to fine-tune the learned representations, will the proposed simple transformation still work? 


\begin{figure}[t]
\setlength{\abovecaptionskip}{-0cm} 
\centering
\centerline{\includegraphics[width=1.0\linewidth]{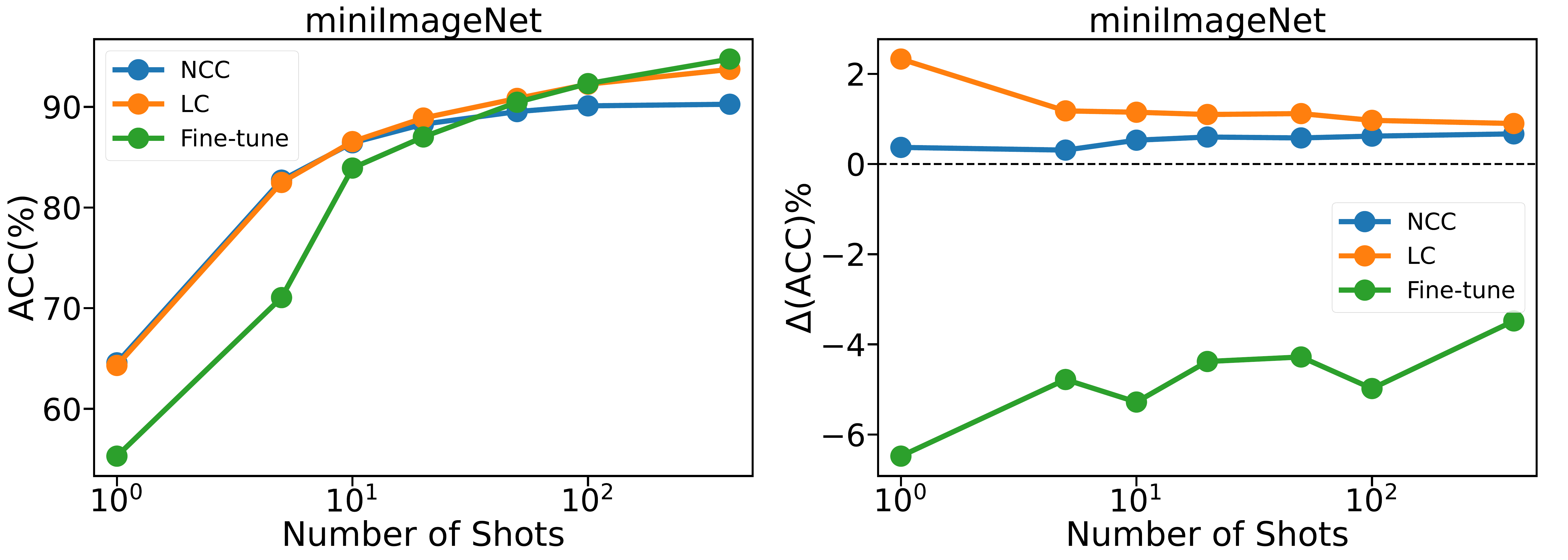}}
\caption{Shot analysis of \emph{mini}ImageNet. Left: performance of different test-time methods. Right: performance gains of the simple transformation using different test-time methods.}
\vskip -0.2in
\label{shot_analysis}
\end{figure}
%

In order to give answers to these questions, we conduct shot analysis experiments on three representative test-time methods that are adopted or are the basis of most mainstream few-shot classification algorithms: (1) The metric-based method Nearest-Centroid Classifier (NCC) presented in ProtoNet, which first average image features of each class in the support set to form class centroids and then assign query features to the class of the nearest centroid; (2) Linear Classifier (LC), which trains a linear layer upon learned image features in the support set, and (3) Fine-tuning, which fine-tunes the feature extractor together with the linear layer using images in the support set. The feature extractor is trained using the state-of-the-art S2M2 algorithm on the training set of \emph{mini}ImageNet, and we test it on the test set of \emph{mini}ImageNet using the above three test-time methods with different numbers of labeled images in each class of the support set. The results are shown in Figure \ref{shot_analysis}. We show the original accuracy of all methods, as well as the impact of simple transformation on the performance.

We first take a look at the right plot, which shows the impact of the simple transformation on all the methods. The performance gains on NCC and LC stay at a relatively high value for all tested shots, which is up to 400 labeled images per class. This indicates that the channel bias problem is not only linked to few-shot settings, but also exists in many-shot settings. However, when we have abundant support images, we
have an alternative choice of fine-tuning the feature extractor directly. Fine-tuning methods have the potential to fully resolve the channel bias problem by directly modifying the image representation and rectifying the channel distribution. The right figure shows that the simple transformation does not improve fine-tuning methods, so indeed the channel bias problem has been largely alleviated. In the left figure, the fine-tuning method exhibits its advantages in many-shot setting, but falls short in few-shot settings. Therefore, we can infer that the channel bias problem exists only in the few-shot setting where freezing the feature extractor and building the classifier on learned features becomes a better choice.

We also have another notable observation. While the performance gain of simple transformation on NCC stays around a fixed value, the performance gain on LC decreases with the increase of shots. Thus the channel bias problem is alleviated to some extent in many-shot settings. This is because more labeled data tells the linear classifier
sufficient information about intra-class variance of data,
making it possible to adjust MMC by modifying the
scale of each row of the linear transformation matrix. So Linear Classifier
can stably increase its performance when more labeled data comes in, until no more linear separation can be achieved, and also the time fine-tuning should get into play to adjust the feature space directly.

\begin{figure*}[!t]
\setlength{\abovecaptionskip}{-0cm} 
\centering
    \subfigure[Singular Value Spectrum] {
    \includegraphics[width=0.24\linewidth]{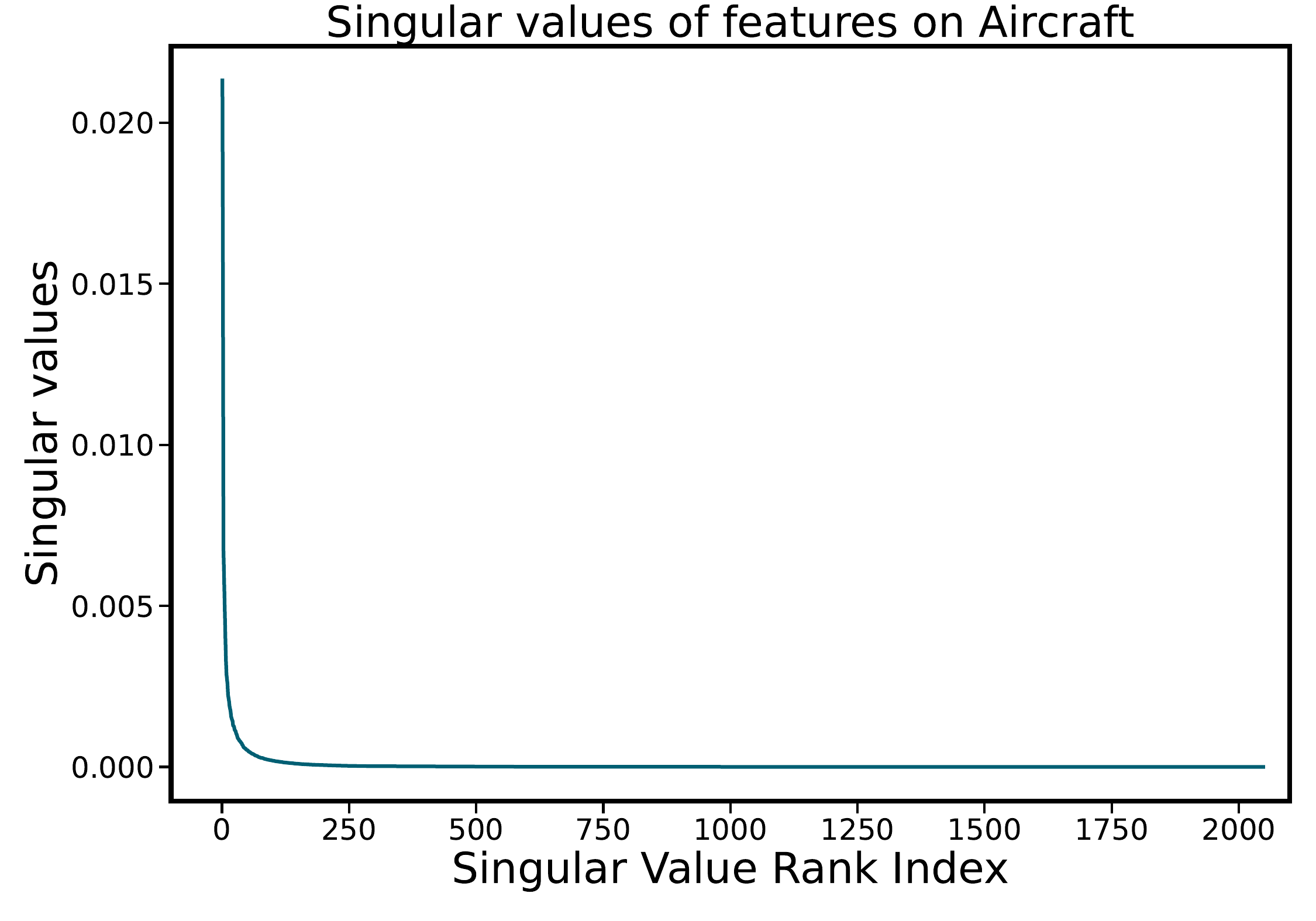}}
    \subfigure[Linear Probing Redundancy] {
    \includegraphics[width=0.24\linewidth]{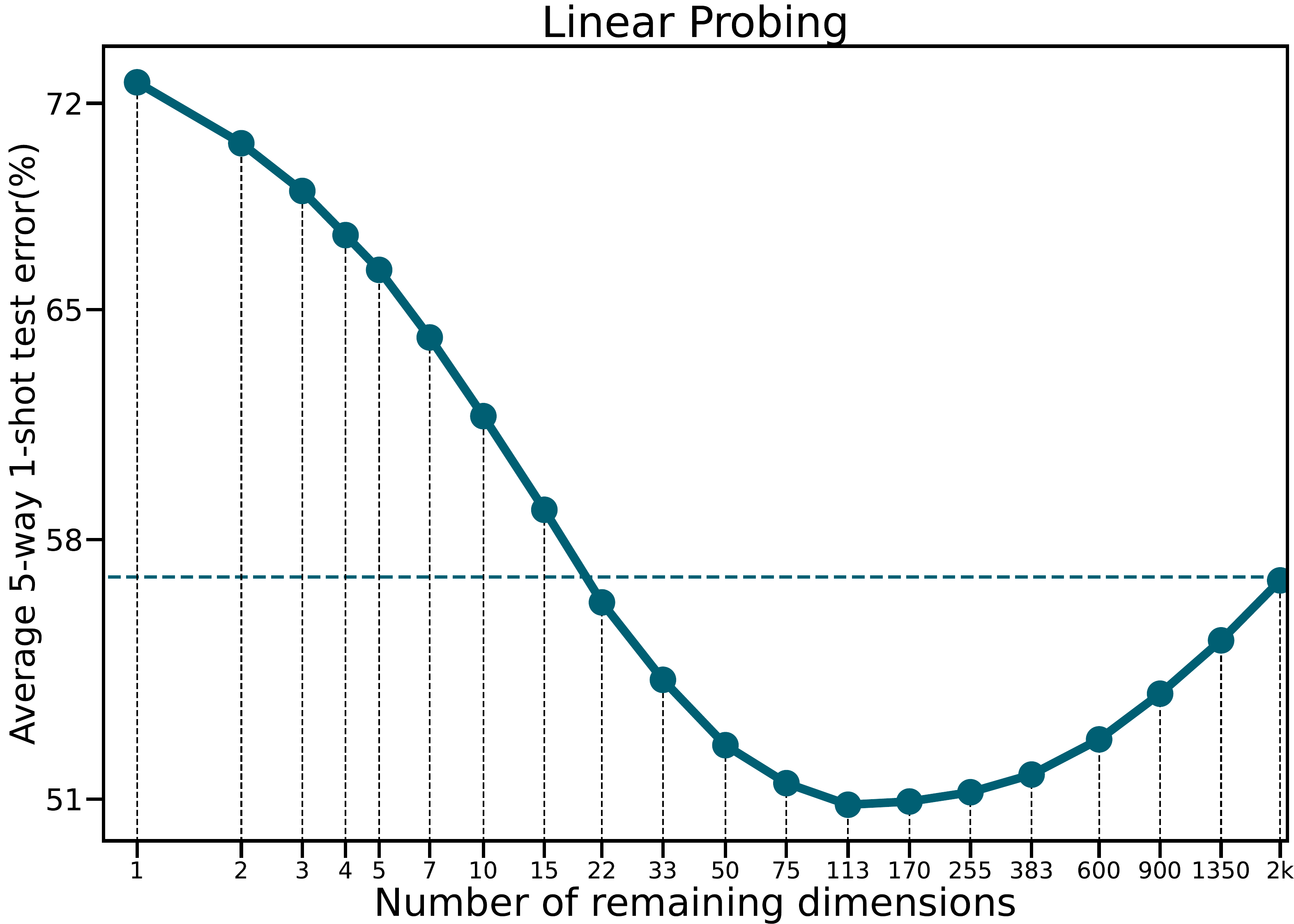}}
    \subfigure[Finetune Redundancy] {
    \includegraphics[width=0.24\linewidth]{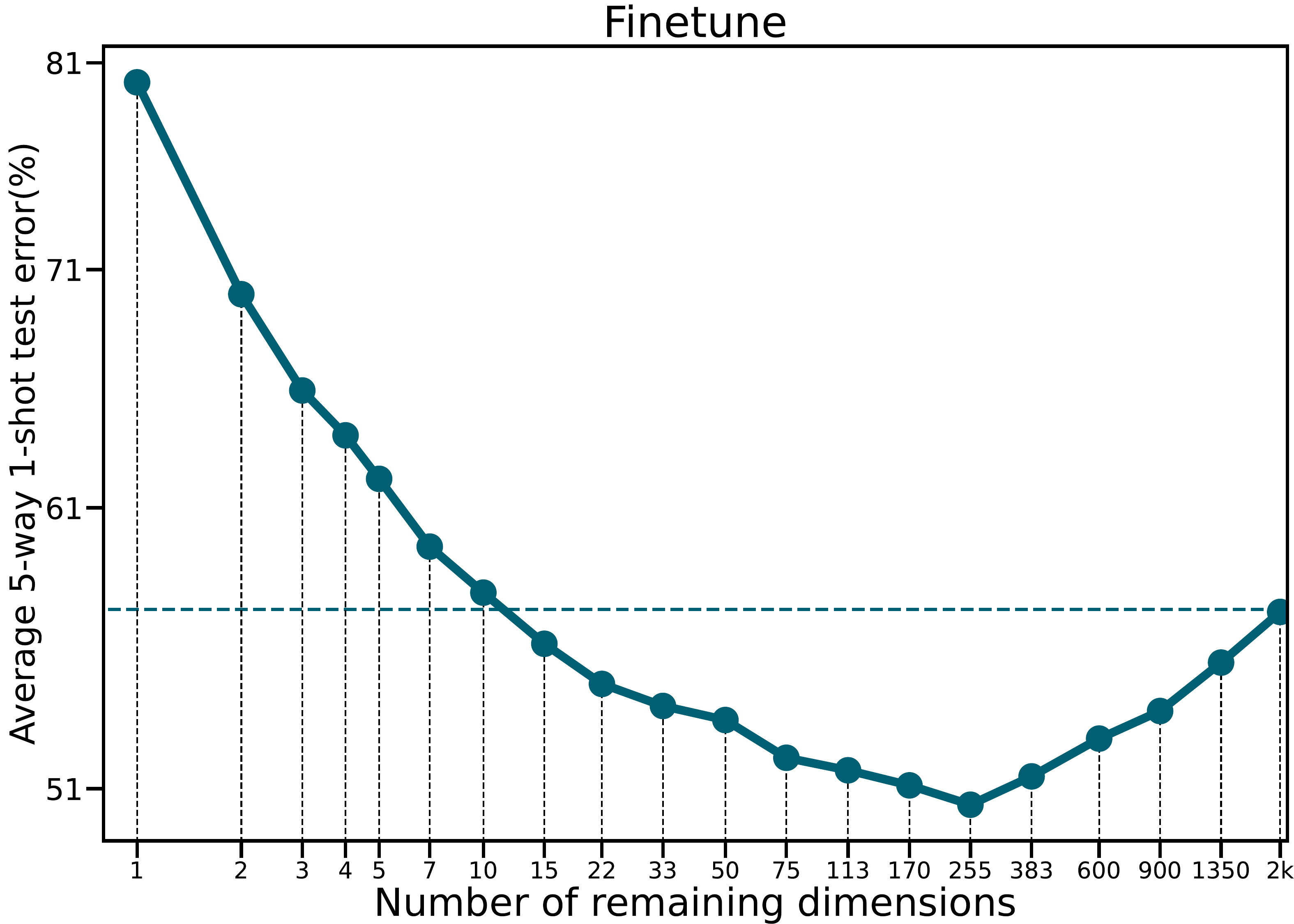}}
    \subfigure[Importance Distribution] {
    \includegraphics[width=0.24\linewidth]{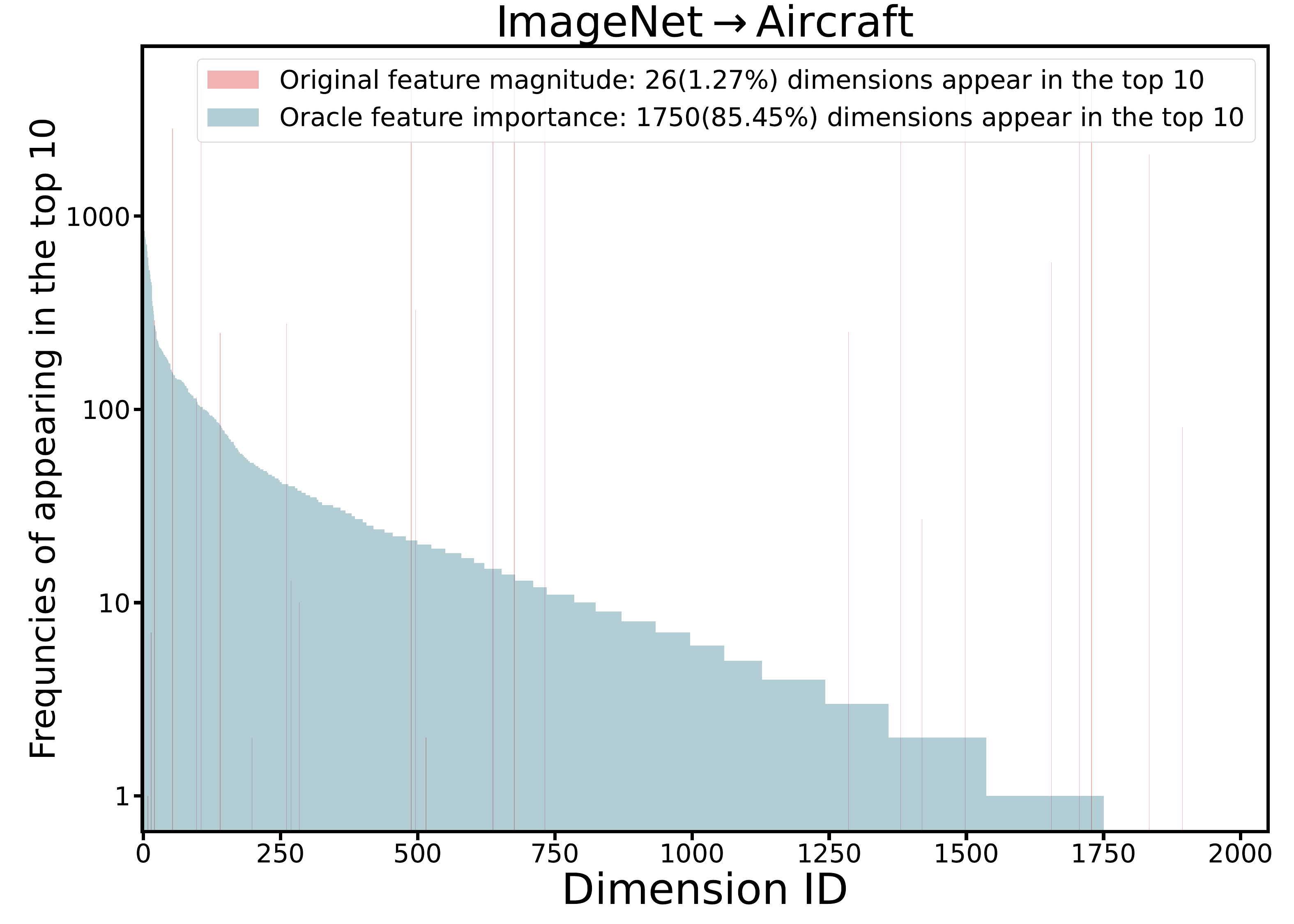}}
\caption{Feature redundancy phenomenon of pre-trained models. (a) Singular value spectrum of the ImageNet-pre-trained ResNet-50 feature space on Aircraft, showing a low effective rank. (b, c) Features are highly redundant for few-shot transfer. Masking unimportant dimensions improves performance for both linear probing and finetuning. Test error is averaged over 2000 5-way 1-shot tasks from Aircraft. (d) Frequencies of feature dimensions appearing in the top 10 most important (blue) vs. top 10 largest magnitude (red) across all binary tasks from Aircraft, showing a mismatch between what is important and what the model emphasizes.}
\label{fig:feature_redundancy}
\end{figure*}

\begin{figure*}[!t]
\setlength{\abovecaptionskip}{-0cm} 
\centering
\includegraphics[width=\textwidth]{Images/allwayshot_maskonly_dimension_analysis_aircraft.pdf}
\caption{How the feature redundancy phenomenon changes with the number of ways and shots. The red curves show the performance obtained by first masking out unimportant features and then adjusting the feature importance of the remaining dimensions via the oracle. The performance of simply masking is shown in green. We use Traffic Signs as the downstream dataset for the second row to have enough shots per class.}
\label{fig:allwayshot}
\end{figure*}

\begin{figure*}[t]
\setlength{\abovecaptionskip}{-0cm} 
\centering
\centerline{\includegraphics[width=1\linewidth]{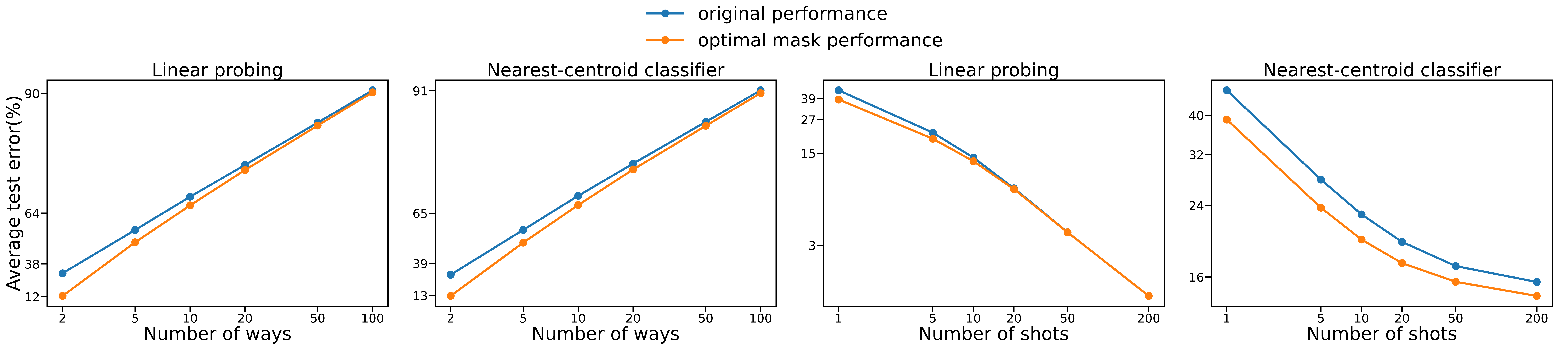}}
\caption{The comparison between the original test error and the error obtained by using the optimal mask when we vary the number of ways and shots. Both axes are logit-scaled. Best viewed in color.}
\label{trend}
\end{figure*}

\section{A Deeper Dive: The Feature Redundancy Phenomenon}
The channel bias problem has an even more dramatic consequence in the few-shot setting. While channel bias suggests that feature importances are wrong, we find that for many tasks, the vast majority of feature dimensions are not just wrongly weighted but are entirely detrimental. For this investigation, we focus our analysis on the Linear Probing (LR) and Nearest-Centroid Classifier (NCC) adaptation methods. This choice is deliberate; extensive empirical studies have shown that the channel bias phenomenon is general across a wide variety of training algorithms \cite{luo2022feature}, and that LR and NCC are not only simple and efficient but also surprisingly strong baselines that are representative of a large class of few-shot adaptation strategies \cite{chen2019closer,tian2020rethinking,luo2023closer}.

\subsection{Feature Redundancy of Pre-trained Models.}
Figure \ref{fig:feature_redundancy}(a) shows the singular value decomposition in sorted order on the correlation matrix of pretrained features collected on the Aircraft Dataset \cite{Aircraft}. As seen, most singular values are very close to zero. In fact, the effective rank \cite{roy2007effective} of the feature space (see details in \textbf{Appendix G} is $86.40$, only about $4.3\%$ of the feature dimensionality $2048$. This means that the feature dimensions are highly repetitive and redundant: we can find a very small feature subspace that is equivalent to the original feature space. Are all information contained in this subspace useful for downstream task performance? We select useful dimensions by utilizing feature importance calculated by Eq. \ref{oracle_adjustment}. As there may be more than two classes in a practical downstream task, we first generalize Eq. \ref{oracle_adjustment} to make it applicable to multi-class scenarios, simply by averaging the feature importance of all binary tasks constructed by each pair of classes in the downstream task. For a given downstream task, we rank feature dimensions according to their feature importance wrt the task, and then gradually decrease the information contained in the features by masking out unimportant dimensions. We show the 5-way 1-shot performance In Figure \ref{fig:feature_redundancy}(b) and \ref{fig:feature_redundancy}(c). We can observe that, as we remove feature dimensions gradually (from right to left), few-shot transfer error gradually decreases. This trend continues until the number of remaining dimensions reaches around 110 for linear probing and 255 for finetuning, which is about only 5-12\% of the full dimensions. Notably, only using these useful dimensions decreases test error significantly, and using about 1\% dimensions can still obtain an accuracy comparable to using the full representation, showing a strong phenomenon of feature redundancy.
Please see \textbf{Appendix H} for experiments on more pretrained
models and downstream datasets.

\textbf{Remark.} The existence of feature redundancy phenomenon indicates that, on the feature level, there exists a notable mismatch between knowledge residing in the pretrained model and the knowledge needed for downstream tasks, at least under few-shot settings. Those feature dimensions that are particularly useful for the task can be seen as \emph{task-specific features} that specialize in the skill needed for the task. This point of view is similar to the observation in \cite{TaskLocalization}, where they found that there exists a small set of \emph{task-specific} parameters ($<1$\% of
model parameters) responsible for the downstream performance of pretrained models, in the sense that only fine-tuning these parameters gives a performance almost as well as the fully fine-tuned model. The difference is that, task-specific features represent the part of pretrained knowledge that is useful for the current task, while task-specific parameters represent new task knowledge the pretrained model currently lacks and should be learned through fine-tuning. The feature redundancy phenomenon also highlights the importance of feature selection for few-shot transfer.

\textbf{Different tasks have different task-specific features.} We now see the importance of task-specific features for downstream few-shot transfer tasks. Then do different tasks have different task-specific features? We calculate the feature importance of all possible binary tasks from Aircraft ($n$ classes produce $n(n-1)/2$ tasks), and visualize the number of times each feature dimension shows up among the top 10 most important dimensions in Figure \ref{fig:feature_redundancy}(d). Surprisingly, more than 85\% of dimensions have been among the most important features in at least one task. Such diversity of discriminative features is unexpected since all tasks are sampled from Aircraft---a fine-grained dataset that seems to, intuitively, have only a small set of discriminative features fixed for all tasks. As a comparison,
we  visualize the number of times the average feature magnitude for each feature dimension ranks in the top 10 across all feature dimensions in Figure \ref{fig:feature_redundancy}(d). Only about 1\% feature dimensions have a top-10 large average magnitude in at least one task, and most of these dimensions have low feature importance in most tasks, showing a mismatch between what the pretrained models focus on and what should be focused on.

\textbf{Downstream dataset size.} Until now, we have fixed the tasks all to be 5-way 1-shot. Will the feature redundancy phenomenon still exist if we increase the downstream dataset size? We show the full picture of how the feature redundancy phenomenon changes with the number of ways or shots for linear probing in Figure \ref{fig:allwayshot}, and give the overall trend for linear probing and NCC in Figure \ref{trend}. In addition to giving results of purely masking unimportant features, we also give the results of first masking unimportant features and then adjusting the magnitude of the remaining feature dimensions in Figure \ref{fig:allwayshot}. We first note that even when we properly adjust feature magnitude (red curves), there are still redundant dimensions harmful to downstream tasks. Thus these dimensions are truly redundant and need to be removed. As seen from the first row of Figure \ref{fig:allwayshot} and the left two plots in Figure \ref{trend}, the feature redundancy phenomenon weakens (but does not disappear) as we increase the number of ways. We have a trivial explanation for this from the previous observations in Figure \ref{fig:feature_redundancy}(d): as different pairs of classes have different important dimensions, more downstream classes turn more dimensions to be important for the discrimination among some classes ($n$ classes $\rightarrow$ $n+1$ classes introduce $n$ additional pairs), hence less severe feature redundancy phenomenon. 






\begin{table}[t]
\setlength{\abovecaptionskip}{0.01cm}  
\scriptsize
\centering
\setlength\tabcolsep{5pt}
\caption{Evaluation results for the two-dimensional Gaussian example.}
\begin{tabular}{cccc}
\hline
& Linear probing/NCC & Linear probing  & NCC\\
& 1-shot & 500-shot & 500-shot\\
\hline
One dimension & \textbf{90.83}$\pm$0.37 & 95.20$\pm$0.010 & \textbf{95.21}$\pm$0.01\\
Two dimensions & 75.87$\pm$0.68 & \textbf{97.36}$\pm$0.007 & 84.37$\pm$0.02\\
\hline
\end{tabular}
\label{toy_example_table}
\end{table}

Now we turn to the second row of  Figure \ref{fig:allwayshot} and the right two plots in Figure \ref{trend}. Interestingly, for linear probing, the feature redundancy phenomenon disappears at some point when the number of shots goes to dozens. The case for NCC is different, where the feature redundancy phenomenon weakens at first and then remains almost unchanged after some point around hundreds of shots. What can we inferred from this observation? We note that most current methods for few-shot learning/transfer can in principle work for many-shot transfer problems, or vice versa. They thus do not answer the question: What makes few-shot transfer different from many-shot transfer? Answering this question may help us understand the unique difficulty of few-shot transfer/learning and design problem-specific methods. Our observation reveals that the feature redundancy phenomenon may be the one we need, as it can serve as an indicator for distinguishing few-shot transfer from many-shot transfer and behave differently with different methods, having the potential of serving as a tool to understand the intrinsic scaling properties of downstream algorithms, as we will do in the next section.



\subsection{What Makes Few-shot Transfer Special?}
\label{Sec: theo-analysis}
\begin{figure}[t]
\setlength{\abovecaptionskip}{-0cm} 
\vskip -0.1in
\centering
\centerline{\includegraphics[width=1.0\linewidth]{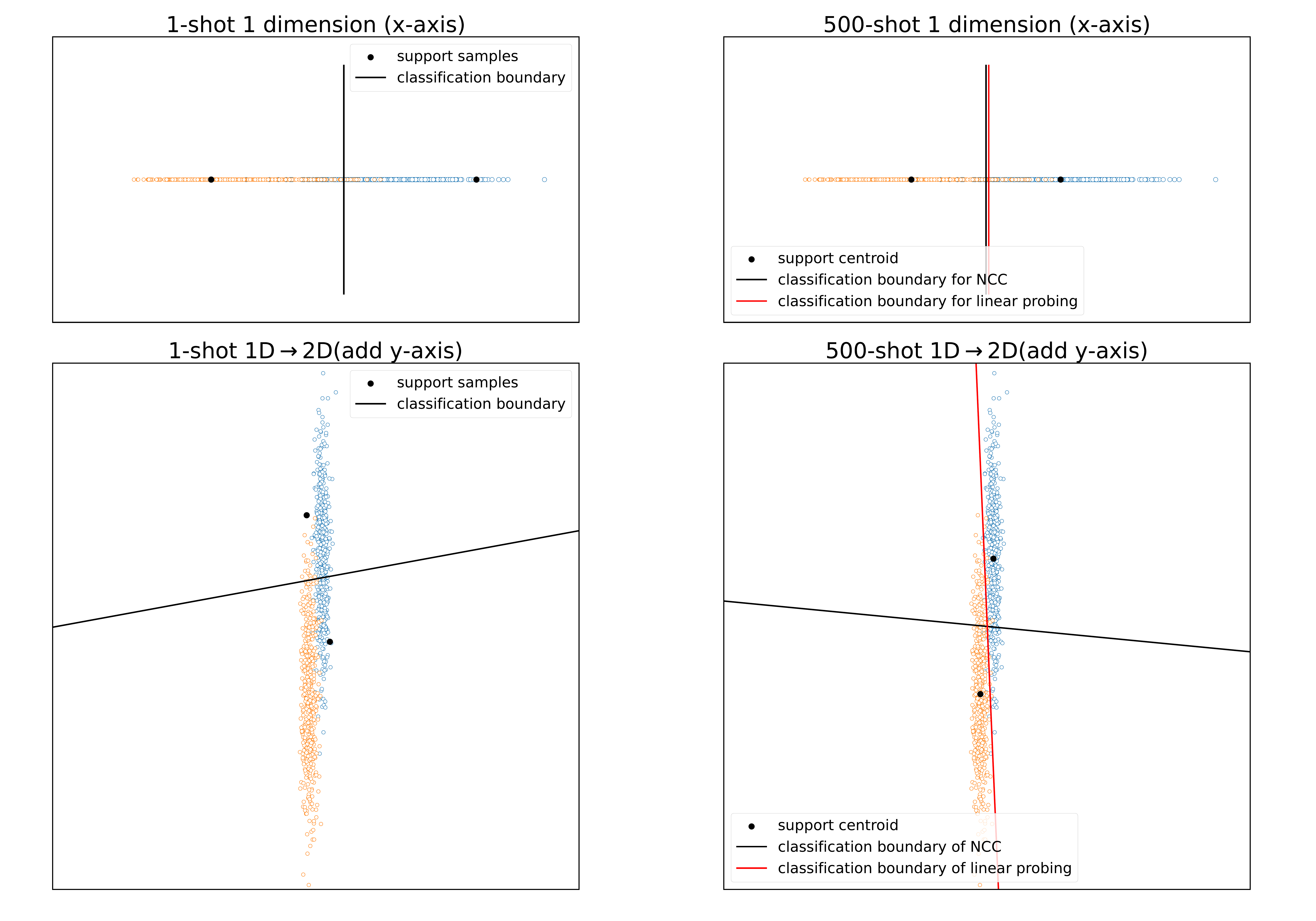}}
\caption{Toy example that illustrates (1) why feature dimensions can be redundant under few-shot settings but turn to be useful when the number of shots increases, and (2) why for NCC the feature redundancy phenomenon does not disappear when the number of shots goes to even hundreds. Best viewed in color.}
\label{gaussian_intuition}
\end{figure}

In this section, we provide some theoretical understandings
that explain several observations from the
previous experiments, including
(1) why the feature redundancy phenomenon exists at low-shot settings, (2) why for linear probing, the feature redundancy phenomenon diminishes quickly when we increase the number of shots, and (3) why for NCC, the feature redundancy phenomenon does not disappear when the number of shots goes to even hundreds. Specifically, we consider a two-dimensional Gaussian model for a binary task that is simple for analysis but can still shed light upon the aforementioned questions for complex high-dimensional cases.  for label $y\in\{a,b\}$, denote the feature as $z_y$, and we consider the model:
\begin{equation}
\label{Gaussian}
    z_y\sim \mathcal{N}\left([\mu_{(y,1)},\mu_{(y,2)}]^T,\mathrm{diag}(\sigma_1^2, \sigma_2^2)\right),
\end{equation}
where $\mu_{(y,i)}$ is the mean vector of the $i$-th dimension for class $y$, and $\sigma_i^2$ is the feature variance for the $i$-th dimension (we assume the same feature variances for the two classes). To give an intuitive understanding of the feature redundancy phenomenon, we instantiate an illustrative example of this model, with $\mu_{(b,1)}=-\mu_{(a,1)}=1$, $\mu_{(b,2)}=-\mu_{(a,2)}=10$, $\sigma_1=0.6$, and $\sigma_2=10$. This distribution is specially designed, with the feature importance $\omega_1=1.67$ of the first dimension much larger than the feature importance $\omega_2=1$ of the second dimension. We report few-shot classification performance averaged over 2000 tasks sampled from this distribution using the first or both dimensions in Table \ref{toy_example_table}, and visualize a 1-shot task and a 500-shot task in Figure \ref{gaussian_intuition}.


\textbf{Empirical intuition.} As seen from Table \ref{toy_example_table} and Figure \ref{gaussian_intuition}, under 1-shot settings, the first dimension is already discriminative enough for classification: both linear probing and NCC achieve 90.83\% accuracy (they are actually the same for 1-shot task). When the second dimension comes in, 1-shot accuracy decreases to 75.87\%, indicating that the new dimension is severely redundant and disturbs the classification. We can observe what leads to this result from the lower left plot in Figure \ref{gaussian_intuition}. As the classifier only sees one training sample per class, the classifier judges a dimension to be informative when the distance between the training samples on this dimension is large. Since for the second dimension, the feature variance $\sigma_2$ is large, the constructed classification boundary is likely to rely more on the second dimension and less on the first dimension. This is problematic because, since the feature variance of the second dimension $\sigma_2$ is large compared to the expected distance between samples $|\mu_{(b,2)}-\mu_{(a,2)}|$ ($\omega_2$ is small), the positions of training samples and test samples on the second dimension will be both highly unstable, which means that the classification boundary will become highly unstable and the test samples are very likely to distribute across the boundary, leading to decreased performance.


From this point of view, it can be easier to understand why in Table \ref{toy_example_table}, NCC still meets the feature redundancy problem with 500 samples per training class, but linear probing does not. As seen from Figure \ref{gaussian_intuition}, we can regard a 500-shot classification using NCC as a 1-shot problem where the two well-estimated class centroids have much smaller variances. This avoids the problem of having a too-large distance between training samples on the y-axis caused by high variance, so we can see that the problem is alleviated to some degree. However, if the difference between the means of each class on the y-axis is considerably large, as in the case of our example, the classification boundary will still be somewhat biased to the y-axis; moreover, although the variance of class centroids is small, the variance of the test samples does not decrease, so they are still very likely to pass through the classification boundary when the feature variance of the second dimension $\sigma_2$ is very large compared to the expected distance between samples $|\mu_{(b,2)}-\mu_{(a,2)}|$ ($\omega_2$ is small). So regardless of the number of shots, NCC will meet feature redundancy problem when the feature importance of the second dimension is considerably smaller than that of the first dimension. 
Unlike NCC, linear probing has a full sense of all training samples and does its best to separate different classes of samples. In addition to estimating means of features, linear probing can estimate the variance of features better as more training data comes in. This leads to an unbiased classifier when the number of training samples goes to infinity. This explains why in Table \ref{toy_example_table} the 500-shot performance using both dimensions is better than that using a single dimension---any dimension can be useful for linear probing under high-shot settings as long as it carries new discriminative information. In our Gaussian data example where the two dimensions are independent, a better-than-chance discriminative ability of a dimension suffices to give new information, and the second dimension clearly satisfies this condition.

\textbf{Theoretical verification.} We give precise theorems for the intuitions shown above. For a task sampled from equation \ref{Gaussian} and $y\in\{a,b\}$, $i\in\{1,2\}$, denote $p_{(y,i)}$ the $i$-th dimension component of the centroid of training samples of class $y$, and  $z_i$ the $i$-th dimension component of a random test sample from class $a$, then we have
\begin{theorem}[Existence of feature redundancy for NCC and 1-shot linear probing]
\label{thm:NCC}
Let $n$ be the number of shots. Suppose that $\frac{|\mu_{(a,2)}-\mu_{(b,2)}|}{\sigma_2}>\frac{2.4}{\sqrt{n}}$ and $\frac{|\mu_{(a,1)}-\mu_{(b,1)}|}{\sigma_1}>2\frac{|\mu_{(a,2)}-\mu_{(b,2)}|}{\sigma_2}+\frac{5.4}{\sqrt{n}}$, then with probability at least $0.9$ over the random draw of the training set, it holds that
\begin{align}
    &\pr\left[\left\|\begin{pmatrix}
 z_1 \\
  z_2\\
\end{pmatrix}-\begin{pmatrix}
 p_{(a,1)} \\
  p_{(a,2)}\\
\end{pmatrix}\right\|_2>\left\|\begin{pmatrix}
 z_1 \\
  z_2\\
\end{pmatrix}-\begin{pmatrix}
 p_{(b,1)} \\
  p_{(b,2)}\\
\end{pmatrix}\right\|_2\right]\nonumber\\
 &>\pr\left[\left|z_1-p_{(a,1)}\right|>\left|z_1-p_{(b,1)}\right|\right].
\end{align}


\end{theorem}
Proofs are given in \textbf{Appendix I}.
Theorem \ref{thm:NCC} basically states that, if we use NCC as the classifier, then regardless of the number of shots, adding a new feature dimension will very likely lead to decreased performance as long as the new dimension has a relatively smaller feature importance defined in equation \ref{oracle_adjustment} compared to other feature dimensions.
The theorem also justifies our dimension ranking criterion based on the magnitude of feature importance. For linear probing, we have



\begin{figure*}[t]
\setlength{\abovecaptionskip}{-0cm} 
\centering
\centerline{\includegraphics[width=\linewidth]{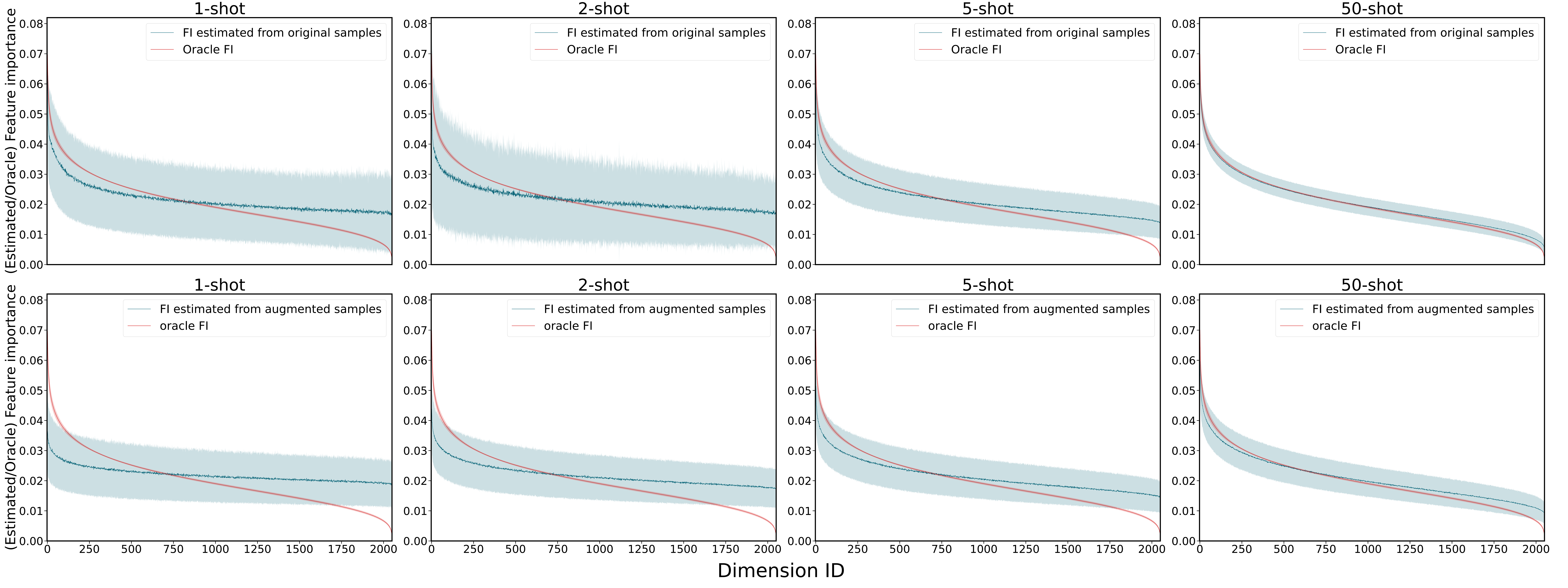}}
\caption{Visualization of how well the feature importance (FI) estimated from few samples approximates the oracle FI. In the first row, FI is estimated using original training samples, while in the second, FI is estimated using samples augmented from the original samples. In each plot, we sample 2000 tasks, and each dimension ID represents the dimension that has the $i$-th largest oracle FI in each task. The curves show the average estimated FI and oracle of all these dimensions, and the bands give standard deviation. Best viewed in color.}
\label{estimate_vs_oracle5way_FI}
\end{figure*}

\begin{figure*}[t]
\setlength{\abovecaptionskip}{-0cm}  
\centering
\centerline{\includegraphics[width=\linewidth]{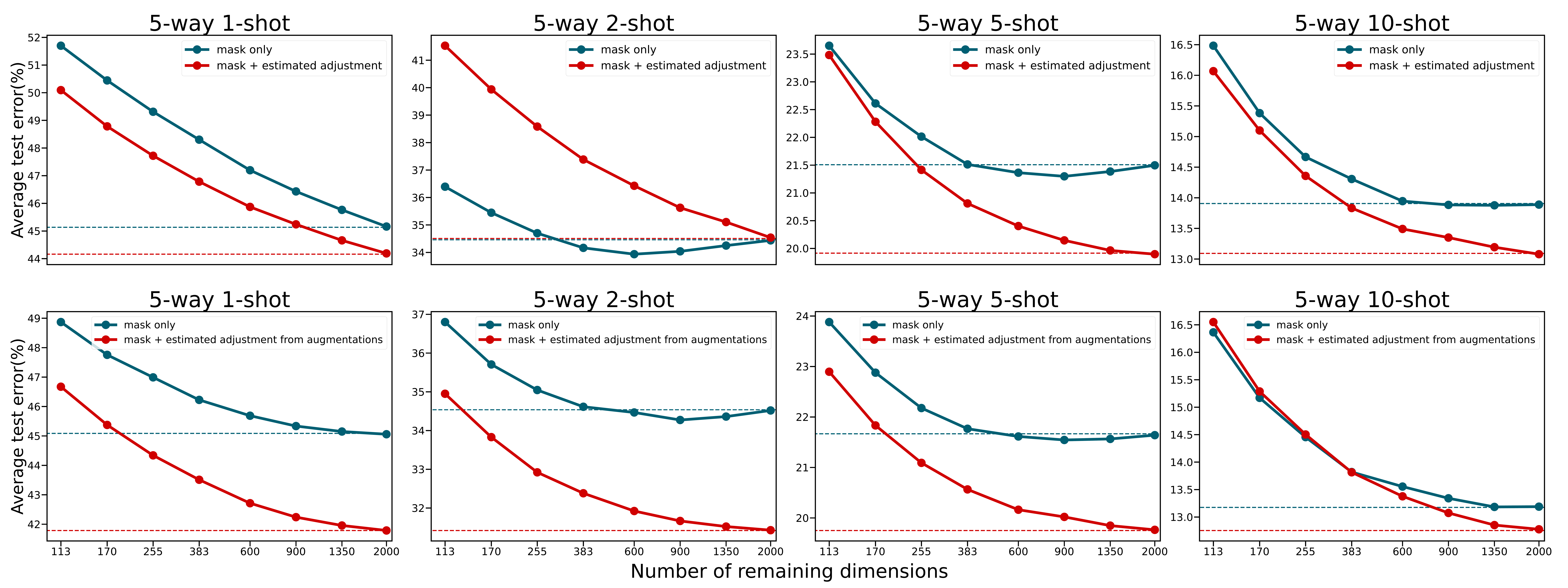}}
\caption{Redundant features cannot be estimated well using few data, but the estimated feature importance can serve as soft masks that help improve few-shot transfer performance. The red curves in the first row show the results of estimation from the original samples, while those in the second row show the results of estimation from augmented samples. Best viewed in color.}
\label{estimate_vs_oracle5way_acc}
\vskip -0.2in
\end{figure*}

\begin{theorem}[Benefit of more feature dimensions for Linear Probing with more shots]\label{thm:linearprob}
Let $n$ be the number of shots, then let $\hat h_{1,n}$ and $\hat h_{2,n}$ be the optimal empirical linear classifiers learned by linear probing using only the first dimension and all dimensions respectively. Suppose that $n>4$, $9\sqrt{\log n/n}<1$ (to avoid trivial bound), and the mean and variance of data are in the constant order, then denote $w_1 = |\mu_{(a,1)}-\mu_{(b,1)}|/2\sigma_1$ and $w_2 = |\mu_{(a,2)}-\mu_{(b,2)}|/2\sigma_2$, we have with probability at least $0.9$ over the random draw of the training set, it holds that
\begin{align*}
&\mathrm{Pr}\big[\hat h_{2,n}([z_1,z_2])\neq y\big]- \mathrm{Pr}\big[\hat h_{1,n}(z_1)\neq y\big] \notag\\
&\hspace{1.8cm}\le \underbrace{\Phi(w_1) - \Phi\Big(\sqrt{w_1^2+w_2^2}\Big)}_{\le 0} + 9\sqrt{\frac{\log n}{n}},
\end{align*}
where $\Phi(\cdot)$ is the cumulative distribution function of the standard Gaussian distribution which is monotonically increasing. 
\end{theorem}
Theorem \ref{thm:linearprob} basically states that, for linear probing, adding a new feature dimension will more likely lead to improved performance when the number of shots $n$ increases, i.e., the upper bound becomes negative. As we show in the proof in \textbf{Appendix J}, the intuition comes from the fact that the optimal Bayes classifier can make the best use of the added dimension as long as it brings new information, and we prove that the optimal empirical linear classifiers will converge to the optimal Bayes classifier as the number of shots goes to infinity, thus letting the newly added dimension more likely to be useful.

\section{Estimating Feature Importance with Few Samples}
Having established the importance of masking redundant features, we turn to the practical question: can we estimate feature importance accurately from only a few samples? The oracle importance in Eq. \eqref{oracle_adjustment} requires global statistics (i.e., true means and variances), which are unavailable in a real few-shot scenario. A natural solution is to leverage the support set to compute sample means and variances as unbiased estimators.

\begin{table*}[t]
\setlength{\abovecaptionskip}{0.01cm} 
\setlength\tabcolsep{3.5pt}
\footnotesize
\caption{5-way few-shot transfer performance improvement brought by adjusting feature magnitude based on the estimation of feature importance from augmented samples, averaged over 2000 tasks using \textbf{linear probing} as the transfer method.}
\label{oracle_performance}
\centering
\begin{tabular}{cc|ccccccccccc}

\hline
& & CUB & Traffic & Aircraft & CropD & ESAT & Fungi & ISIC & Omniglot & QuickD & Flowers & ChestX\\
\hline
\multirow{2}{*}{DINOv2} & 1-shot & \acc{97.00}{0.0} & \acc{51.79}{3.0} & \acc{66.50}{0.0} & \acc{90.08}{0.1} & \acc{64.68}{1.5} & \acc{70.83}{0.5} & \acc{31.06}{0.6} & \acc{81.76}{0.4} & \acc{68.26}{0.0} & \acc{99.79}{0.0} & \acc{22.28}{0.0}\\
& 5-shot & \acc{98.84}{0.0} & \acc{73.17}{3.1} & \acc{74.70}{1.0} & \acc{95.63}{0.3} & \acc{80.42}{2.2} & \acc{86.62}{1.1} & \acc{41.83}{1.8} & \acc{93.70}{1.2} & \acc{86.23}{0.2} & \acc{99.93}{0.0} & \acc{25.17}{0.2}\\
\multirow{2}{*}{EsViT-base} & 1-shot & \acc{68.10}{0.4} & \acc{58.16}{1.3} & \acc{33.10}{0.8} & \acc{79.04}{1.6} & \acc{67.15}{2.2} & \acc{52.70}{2.0} & \acc{31.70}{0.7} & \acc{83.58}{2.0} & \acc{52.54}{1.4} & \acc{82.89}{2.1} & \acc{21.89}{0.1}\\
& 5-shot & \acc{84.86}{1.6} & \acc{76.47}{2.2} & \acc{37.48}{7.2} & \acc{92.59}{0.8} & \acc{85.64}{1.2} & \acc{74.30}{3.0} & \acc{44.31}{1.1} & \acc{93.81}{1.4} & \acc{71.74}{1.8} & \acc{96.58}{0.6} & \acc{24.87}{0.2}\\
\multirow{2}{*}{Res50ImageNet} & 1-shot & \acc{74.93}{1.6} & \acc{54.58}{2.7} & \acc{40.74}{1.9} & \acc{74.39}{2.5} & \macc{69.52}{1.0} & \acc{50.36}{2.7} & \acc{29.58}{1.1} & \acc{80.63}{1.6} & \acc{50.96}{1.2} & \acc{79.17}{3.5} & \acc{21.85}{0.2}\\
& 5-shot & \acc{90.76}{1.2} & \acc{73.68}{4.1} & \acc{57.37}{3.7} & \acc{89.48}{1.8} & \acc{84.71}{0.8} & \acc{71.47}{4.0} & \acc{40.56}{2.3} & \acc{91.45}{2.1} & \acc{68.85}{2.7} & \acc{93.57}{1.6} & \acc{24.75}{0.1}\\
\multirow{2}{*}{CLIP-base} & 1-shot & \acc{85.59}{3.0} & \acc{60.51}{4.9} & \acc{65.69}{1.5} & \acc{77.92}{3.5} & \acc{62.64}{4.6} & \acc{53.78}{4.3} & \acc{30.98}{1.2} & \acc{85.79}{3.1} & \acc{65.50}{2.3} & \acc{92.76}{2.2} & \acc{21.29}{0.2}\\
& 5-shot & \acc{95.97}{0.9} & \acc{78.60}{6.2} & \acc{79.29}{1.9} & \acc{89.93}{3.2} & \acc{77.94}{6.6} & \acc{71.64}{6.3} & \acc{41.74}{3.2} & \acc{94.26}{2.6} & \acc{83.18}{2.0} & \acc{98.74}{0.5} & \acc{22.84}{0.3}\\
\multirow{2}{*}{IBOT-base} & 1-shot & \acc{74.78}{1.5} & \acc{56.41}{1.8} & \acc{35.82}{0.0} & \acc{83.69}{0.1} & \macc{73.11}{0.2} & \acc{56.92}{0.3} & \acc{33.62}{0.4} & \acc{87.32}{1.1} & \macc{56.46}{0.2} & \acc{87.62}{1.3} & \acc{23.23}{0.0}\\
& 5-shot & \acc{91.12}{1.4} & \acc{76.60}{3.0} & \acc{47.23}{5.7} & \acc{94.99}{0.4} & \acc{89.07}{0.3} & \acc{76.72}{1.5} & \acc{47.72}{0.9} & \acc{95.84}{1.2} & \acc{73.95}{1.4} & \acc{97.67}{0.4} & \acc{27.15}{0.3}\\
\multirow{2}{*}{Swin-base} & 1-shot & \acc{77.54}{1.2} & \acc{52.53}{2.0} & \acc{42.50}{0.6} & \acc{79.30}{0.8} & \acc{62.29}{2.8} & \acc{52.98}{1.3} & \acc{29.26}{0.6} & \acc{84.90}{0.5} & \acc{57.93}{0.3} & \acc{80.68}{1.0} & \acc{21.87}{0.2}\\
& 5-shot & \acc{90.32}{0.6} & \acc{73.31}{2.5} & \acc{52.92}{3.0} & \acc{92.50}{0.6} & \acc{80.65}{2.1} & \acc{71.11}{2.3} & \acc{39.93}{1.6} & \acc{94.26}{0.9} & \acc{76.47}{1.0} & \acc{94.10}{0.6} & \acc{24.45}{0.3}\\
\hline

\end{tabular}
\vskip -0.1in
\end{table*}

\begin{table*}[t]
\setlength{\abovecaptionskip}{0.01cm} 
\setlength\tabcolsep{3.5pt}
\footnotesize
\caption{5-way few-shot transfer performance improvement brought by the proposed method when using \textbf{finetuning} as the transfer method.}
\label{finetune}
\centering
\begin{tabular}{cc|ccccccccccc}

\hline
& & CUB & Traffic & Aircraft & CropD & ESAT & Fungi & ISIC & Omniglot & QuickD & Flowers & ChestX\\
\hline

\multirow{2}{*}{DINOv2} & 1-shot & \acc{96.21}{0.2} & \acc{51.40}{4.0} & \acc{67.43}{0.0} & \acc{90.95}{0.0} & \acc{63.40}{1.9} & \acc{71.84}{0.4} & \acc{33.52}{1.7} & \acc{81.73}{0.9} & \acc{68.36}{0.3} & \acc{99.87}{0.0} & \acc{22.29}{0.2}\\
& 5-shot & \acc{98.51}{0.0} & \acc{73.13}{4.0} & \acc{76.01}{1.2} & \acc{95.55}{0.4} & \acc{80.17}{2.7} & \acc{85.36}{1.8} & \acc{45.37}{3.4} & \acc{93.17}{1.5} & \acc{82.95}{1.4} & \acc{99.92}{0.0} & \acc{24.04}{0.2}\\
\multirow{2}{*}{EsViT-base} & 1-shot & \acc{68.44}{0.0} & \acc{57.44}{1.2} & \acc{33.71}{0.4} & \acc{81.40}{1.3} & \acc{68.19}{2.3} & \acc{54.49}{2.2} & \acc{34.77}{1.1} & \acc{83.56}{2.3} & \acc{51.51}{0.4} & \acc{85.27}{1.8} & \macc{22.56}{0.1}\\
& 5-shot & \acc{78.27}{3.4} & \acc{73.61}{4.0} & \acc{31.60}{5.9} & \acc{89.13}{2.4} & \acc{81.08}{3.9} & \acc{60.80}{8.3} & \acc{46.51}{4.7} & \acc{93.05}{1.3} & \acc{65.89}{3.0} & \acc{93.63}{2.5} & \acc{23.51}{0.0}\\
\multirow{2}{*}{Res50ImageNet} & 1-shot & \acc{76.48}{1.6} & \acc{55.33}{2.9} & \acc{43.62}{0.5} & \acc{79.21}{0.8} & \macc{72.43}{0.9} & \acc{50.93}{1.6} & \acc{34.57}{0.8} & \acc{81.11}{1.3} & \acc{54.59}{0.2} & \acc{83.13}{1.0} & \acc{22.57}{0.0}\\
&5-shot & \acc{91.89}{1.1} & \acc{81.11}{3.4} & \acc{59.83}{4.7} & \acc{92.20}{1.1} & \acc{87.69}{0.3} & \acc{74.63}{2.9} & \acc{53.23}{3.7} & \acc{95.15}{2.1} & \acc{73.57}{2.2} & \acc{95.83}{0.8} & \acc{24.76}{0.2}\\
\multirow{2}{*}{CLIP-base} & 1-shot & \acc{86.16}{3.7} & \acc{60.52}{5.1} & \acc{64.16}{2.0} & \acc{79.07}{3.6} & \acc{62.91}{4.4} & \acc{52.25}{4.2} & \acc{34.17}{1.6} & \acc{86.79}{2.9} & \acc{61.87}{4.6} & \acc{91.91}{3.2} & \acc{21.05}{0.4}\\
& 5-shot & \acc{95.08}{1.1} & \acc{76.27}{6.9} & \acc{76.64}{3.3} & \acc{89.60}{3.0} & \acc{75.93}{7.0} & \acc{64.83}{8.6} & \acc{43.92}{6.0} & \acc{93.68}{2.8} & \acc{77.75}{4.9} & \acc{98.41}{0.5} & \acc{22.12}{0.0}\\
\multirow{2}{*}{IBOT-base} & 1-shot & \acc{74.00}{1.6} & \acc{55.87}{2.0} & \acc{37.23}{0.3} & \acc{85.52}{0.1} & \macc{73.69}{0.2} & \acc{57.53}{0.3} & \acc{37.16}{0.8} & \acc{88.07}{1.2} & \macc{57.39}{0.6} & \acc{86.83}{1.9} & \acc{23.71}{0.1}\\
& 5-shot & \acc{89.12}{2.3} & \acc{75.01}{3.9} & \acc{46.33}{5.1} & \acc{95.12}{0.3} & \acc{88.89}{0.7} & \acc{75.01}{1.7} & \acc{55.32}{1.2} & \acc{94.84}{1.1} & \acc{73.00}{1.8} & \acc{97.77}{0.3} & \acc{26.88}{0.6}\\
\multirow{2}{*}{Swin-base} & 1-shot & \acc{76.28}{1.5} & \acc{51.69}{1.8} & \acc{42.79}{0.9} & \acc{80.61}{1.1} & \acc{62.23}{3.8} & \acc{50.32}{2.8} & \acc{31.41}{0.8} & \acc{84.13}{0.2} & \acc{55.04}{1.0} & \acc{80.69}{1.3} & \acc{22.47}{0.0}\\
& 5-shot & \acc{85.93}{2.0} & \acc{69.21}{3.4} & \acc{44.29}{4.4} & \acc{89.44}{2.0} & \acc{77.88}{3.1} & \acc{58.47}{7.3} & \acc{41.11}{4.6} & \acc{92.87}{1.2} & \acc{69.77}{3.1} & \acc{89.59}{2.7} & \acc{23.65}{0.8}\\
\hline

\end{tabular}
\vspace {-0.1in}
\end{table*}

\subsection{Difficulty of Estimating Feature Importance}
We show in the first row of Figure \ref{estimate_vs_oracle5way_FI} how well the estimated FI approximates the oracle FI with different numbers of shots. In each plot, the $i$-th dimension ID represents the dimension that has the $i$-th largest oracle FI in each of 2000 sampled tasks. We average all estimated and oracle FIs of these dimensions in each task, and show the results with standard deviation. The distance of the blue lines to the red lines and the variance of two lines in each plot measures how well the estimated FI approximates the oracle FI. We can first observe that, there are a small number of dimensions for each task that have a very large or small oracle FI. As we have analyzed in the previous sections, these dimensions are particularly important: those dimensions having too small oracle FIs are redundant dimensions that need to be masked out, while those having large oracle FIs are task-specific dimensions that need to be highlighted. 
As seen, the estimated FI has a large variance throughout all dimensions under few-shot setting. In particular, we see that the average estimated FI has a large gap with the oracle FI at the head and tail of the dimensions. We blame this phenomenon as the result of the difficulty of determining a small set of special dimensions with a limited number of samples for estimation. This shows the fundamental difficulty of figuring out redundant and task-specific dimensions.
We verify this observation in Figure \ref{estimate_vs_oracle5way_acc}, where we show that using the estimated FIs cannot help us figure out the redundant dimensions. Another interesting observation is that the variance of the estimated FIs for 2-shot setting is even larger than that for 1-shot setting.

\subsection{Improving Estimation with Data Augmentation}
Since the core problem is data scarcity, a natural solution is data augmentation. We generate 5 augmented views (using random cropping) for each support image and use this larger set to estimate the FIs. The second row of Figure \ref{estimate_vs_oracle5way_FI} shows that this reduces the estimation variance. Consequently, using these improved FIs for feature adjustment yields much better performance.

Even with augmentation, hard masking is still challenging because estimation error persists. However, instead of hard masking, we can use the estimated importance vector $\hat{\oo}$ to perform a \emph{soft mask} or feature adjustment, by scaling the features with $\hat{\oo}$. This serves as a robust surrogate: instead of making a hard decision about which features to discard, we down-weight the features we are less confident about. Table \ref{oracle_performance} shows that this simple method, which we call \textbf{A}ugmented \textbf{F}eature \textbf{I}mportance \textbf{A}djustment (AFIA), provides consistent and significant performance gains for linear probing across a wide range of state-of-the-art pre-trained models and downstream datasets.
Crucially, AFIA is not limited to linear probing. As shown in Table \ref{finetune}, the same technique provides significant gains when full finetuning is used as the transfer method. Although finetuning modifies the feature extractor itself, the initial features are still subject to channel bias. We speculate that using AFIA to adjust features at the start of the finetuning process provides a better-initialized representation, guiding the optimization towards a better solution, which is particularly important in data-scarce scenarios. This demonstrates the broad applicability of our proposed method.
We give detailed ablation studies of the method in \textbf{Appendix K}.



\section{Discussion and Related Work}
This work connects several lines of research in FSL as well as representation learning.

\textbf{Task Distribution Shift in FSL.} 
The challenge of distribution shift in FSL is well-recognized. Initial benchmarks like \emph{mini}ImageNet focused only on category shift \cite{vinyals2016matching}. Later work introduced more complex shifts, including domain shifts in cross-domain FSL \cite{chen2019closer,guo2020broader} and granularity shifts in coarse-to-fine FSL \cite{luo2021boosting,bukchin2021finegrained}. Our work demonstrates that the channel bias and feature redundancy problems are common across all three types of shift. Prior work has sought to address distribution shift by adjusting spatial attention \cite{doersch2020crosstransformers} or removing background information \cite{luo2021rectifying}. Our work provides an orthogonal perspective, focusing on the importance of features rather than spatial locations.

\textbf{Test-Time Methods for FSL.} 
Mainstream FSL algorithms can be categorized by their test-time adaptation strategy. (1) \textbf{Optimization-based/Finetuning} methods, like MAML \cite{finn2017modelagnostic}, update model parameters on the support set. While powerful with enough data \cite{kolesnikov2020big}, they often overfit in few-shot scenarios, as shown in our analysis. (2) \textbf{Metric-based} methods, like Prototypical Networks \cite{snell2017prototypical}, use a fixed feature space and a distance metric (e.g., Euclidean or cosine) for classification. Their strong inductive bias is effective in low-data regimes, but as we have shown, they are susceptible to channel bias. (3) \textbf{Linear Classifier (LC)} methods train a linear layer on top of frozen features \cite{chen2019closer,tian2020rethinking}. Our analysis reveals that LC methods can learn to partially mitigate channel bias with more data, bridging the gap between metric-based and finetuning methods.

\textbf{Feature Transformation and Selection in FSL.} 
Several works have explored feature transformation in FSL. Some introduce learnable transformations during training \cite{tseng2020crossdomain} or use generative models to rectify feature distributions \cite{xu2021exploring}. Others adapt normalization layer parameters at test time, like FiLM \cite{perez2018film}, which can be seen as a form of lightweight finetuning \cite{oreshkin2018tadam,zhang2025reliable,wu2025skip,zhang2024dept}. The most related work may be ConFeSS \cite{das2022confess}, which masks task-irrelevant features at test time. Our work provides a theoretical foundation for why such feature selection is beneficial, framing it as a solution to the channel bias and feature redundancy problems. While they focus on hard masking, we show that a softer, magnitude-based adjustment is more robust when importance cannot be estimated perfectly.

\section{Conclusion}
This paper provides a unified and in-depth analysis of a critical challenge in few-shot learning: the inability of pre-trained models to correctly adapt their feature representations to novel tasks. We began by identifying the \textit{channel bias} problem, where models are overconfident in feature importance learned on the source domain. We then showed this leads to \textit{feature redundancy} in the few-shot setting, a phenomenon that we argue is a defining characteristic of few-shot transfer. Our theoretical and empirical analyses explain why this redundancy arises from high-variance, confounding feature dimensions and why it diminishes as more data becomes available, particularly for flexible classifiers like linear probes. While accurately identifying redundant features from a few samples is difficult, we proposed a practical and effective method, AFIA, that uses augmented data to estimate feature importance and apply a soft adjustment to the feature representation. This simple, test-time approach yields consistent improvements across a wide variety of models and benchmarks, for both linear probing and full finetuning. This work deepens our understanding of representation transfer, highlighting that for few-shot learning, less is often more. Future research could explore more sophisticated methods for estimating and leveraging feature importance, or investigate how these phenomena manifest in different adaptation schemes. Ultimately, we hope the insights into channel bias and feature redundancy will inspire the development of more robust and data-efficient learning algorithms.

\appendices
% ...

% ...

% ...




\ifCLASSOPTIONcaptionsoff
  \newpage
\fi

\bibliographystyle{IEEEtran}
\bibliography{ref}

\begin{IEEEbiography}[{\includegraphics[width=1in,height=1.25in,clip]{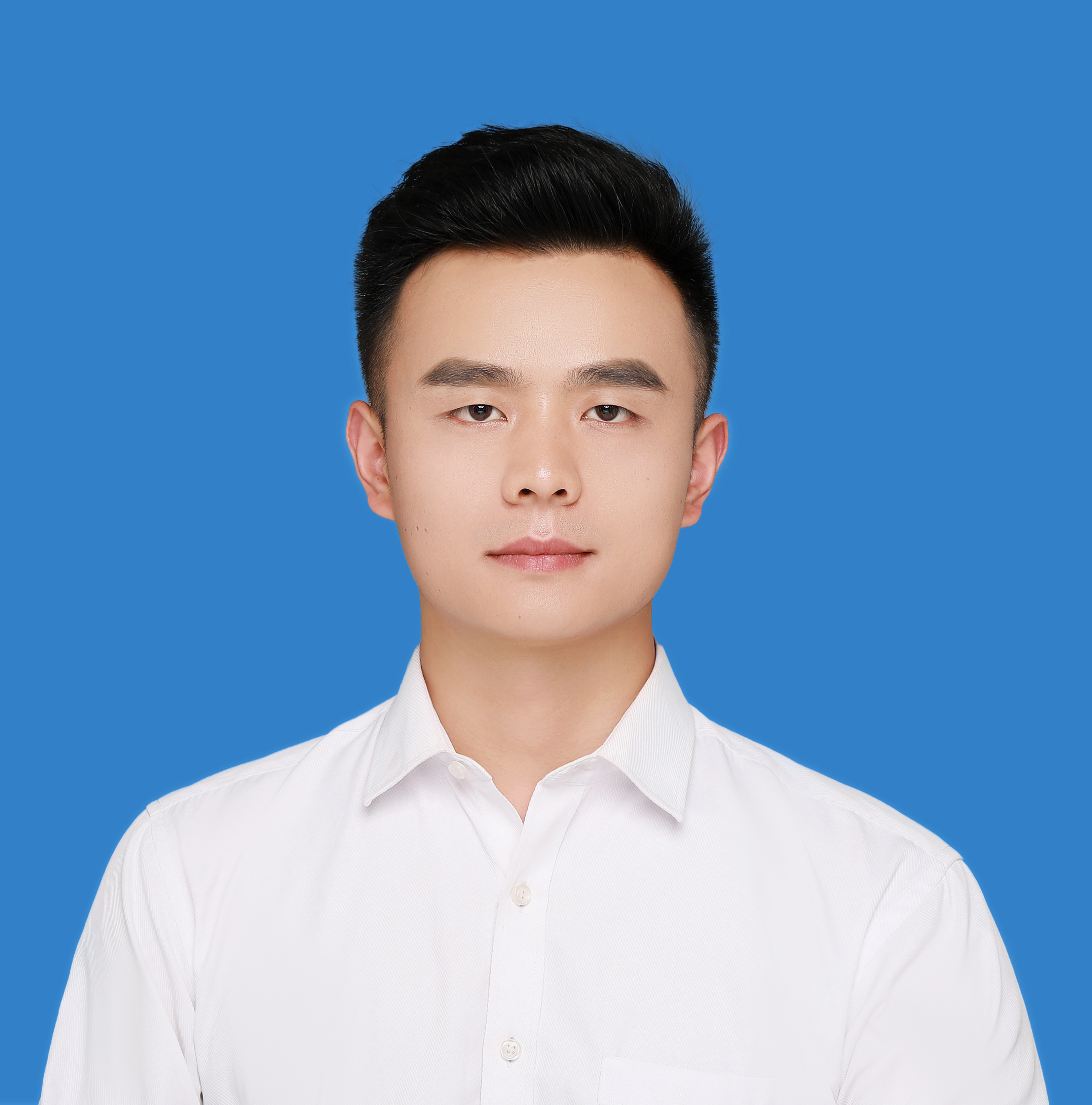}}]{Ji Zhang} is an Assistant Professor with the School of Computing and Artificial Intelligence, Southwest Jiaotong University, China. He obtained his PhD degree from University of Electronic Science and Technology of China in 2024, under the supervision of Prof. Jingkuan Song. His research interests include few-shot learning, transfer learning and robotics. He has published over ten papers on top conferences/journals, such as TPAMI, TIP, CVPR, ICCV, ICML. 
\end{IEEEbiography}

\begin{IEEEbiography}[{\includegraphics[width=1in,height=1.25in,clip]{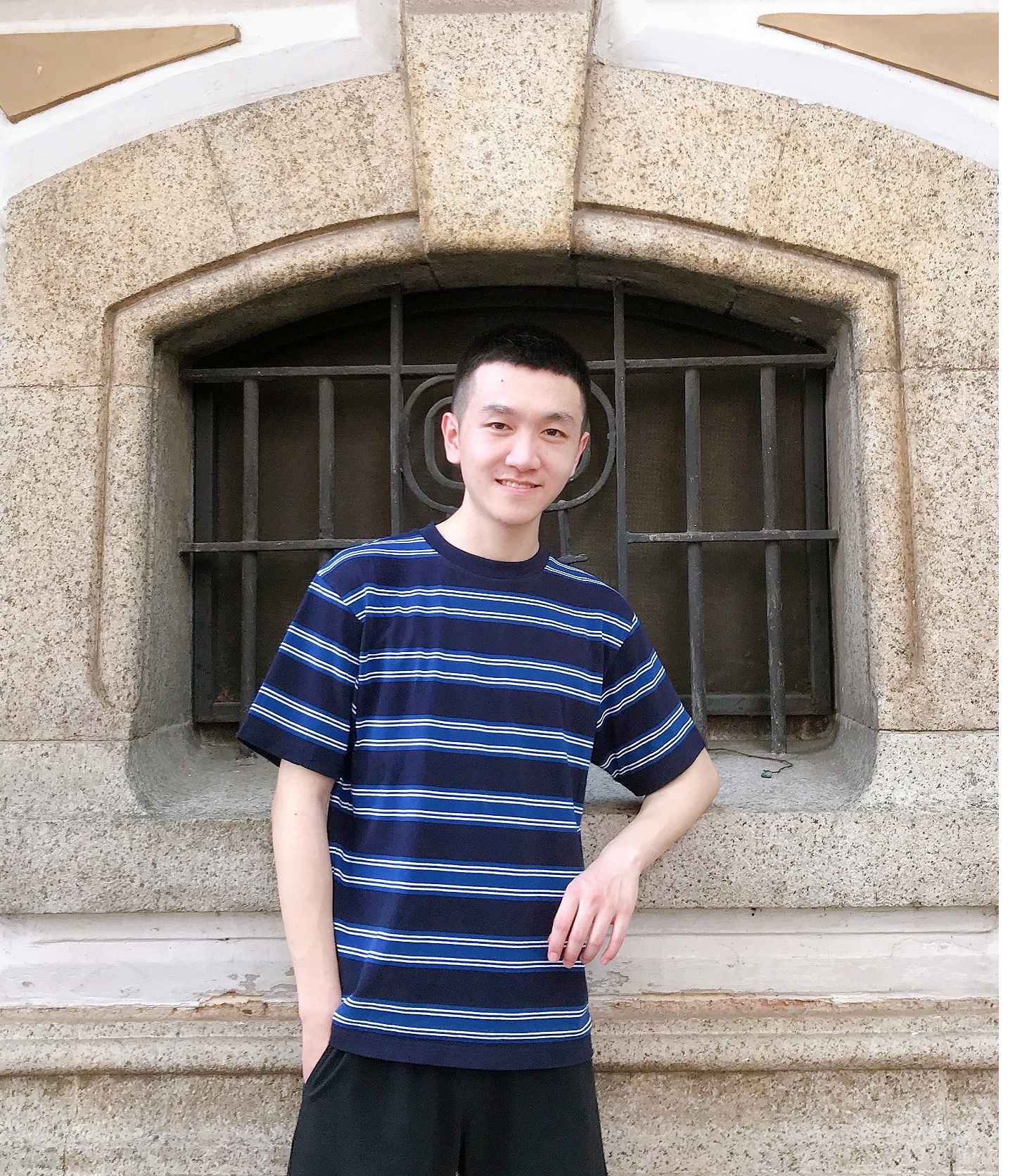}}]{Xu Luo} is a Ph.D. candidate at University of Electronic Science and Technology of China, advised by Prof. Jingkuan Song. His research interests include Computer Vision, LLMs, and Robotics. 
He has published multiple papers on top-tier machine learning conferences including NeurIPS, ICML, ICLR, CVPR, ICCV, CoRL, etc. 
\end{IEEEbiography}

\begin{IEEEbiography}[{\includegraphics[width=1in,height=1.25in,clip]{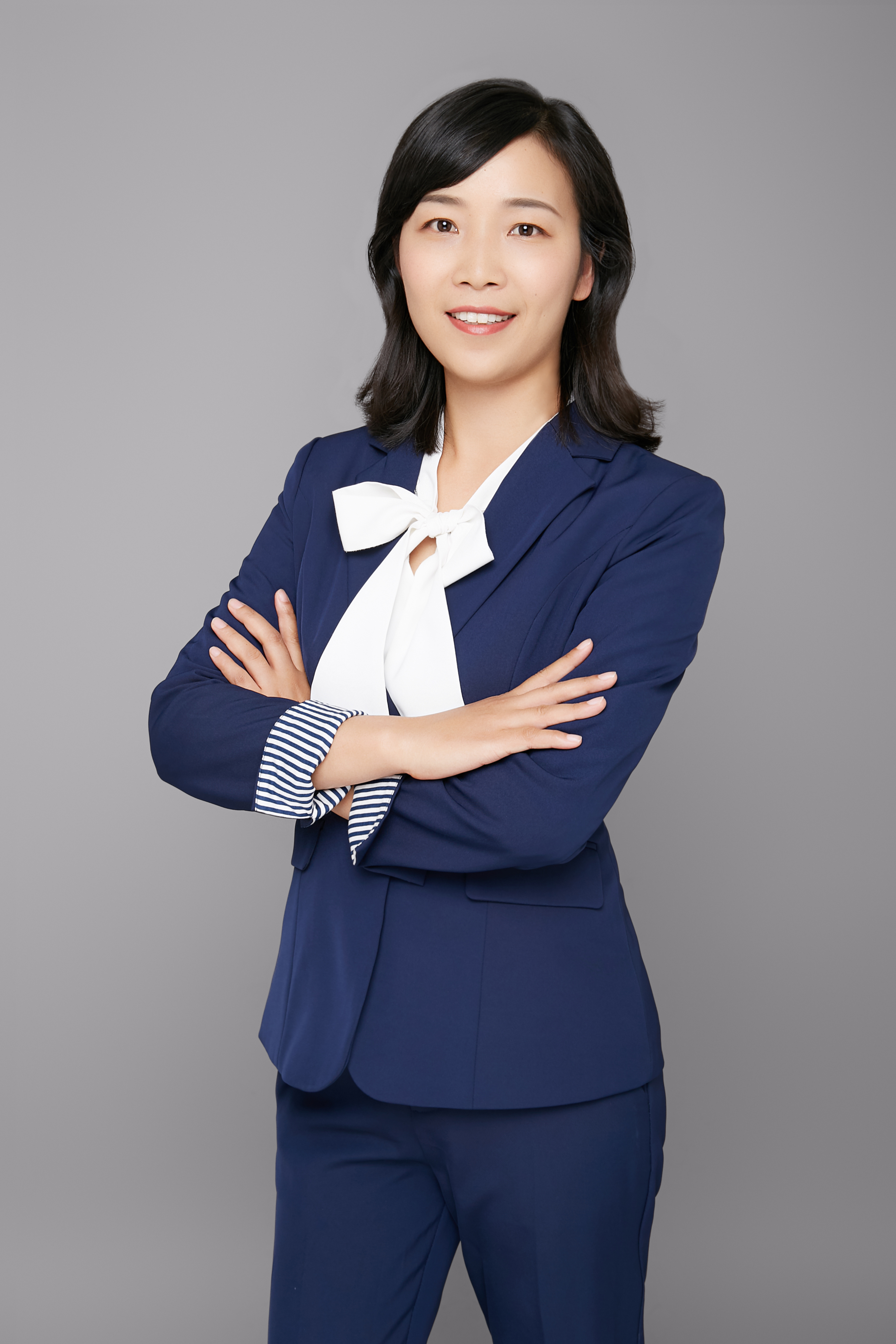}}]{Lianli Gao} is a professor with the School of Computer Science and Engineering, University of Electronic Science and Technology of China. She obtained her PhD degree in Information Technology from The University of Queensland (UQ), Australia, under the supervision of Prof. Jane Hunter and Prof. Michael Bruenig. Her research ranges from Semantic Web, Machine Learning, Deep Learning, Computer Vision (Images and Videos), NLP, Knowledge Reasoning, Knowledge and the related practical applications etc. Specifically, she is mainly focusing on integrating Natural Language for Visual Content Understanding. She has  the winner of the IEEE Transactions on Multimedia 2020 Prize Paper Award, Best Student Paper Award in Australian Database Conference (2017, Australia), IEEE TCMC Rising Star Award 2020 and ALIBABA Academic Young Fellow. She is an Associate Editor of IEEE TMM.
\end{IEEEbiography}

\begin{IEEEbiography}[{\includegraphics[width=1in,height=1.25in,clip]{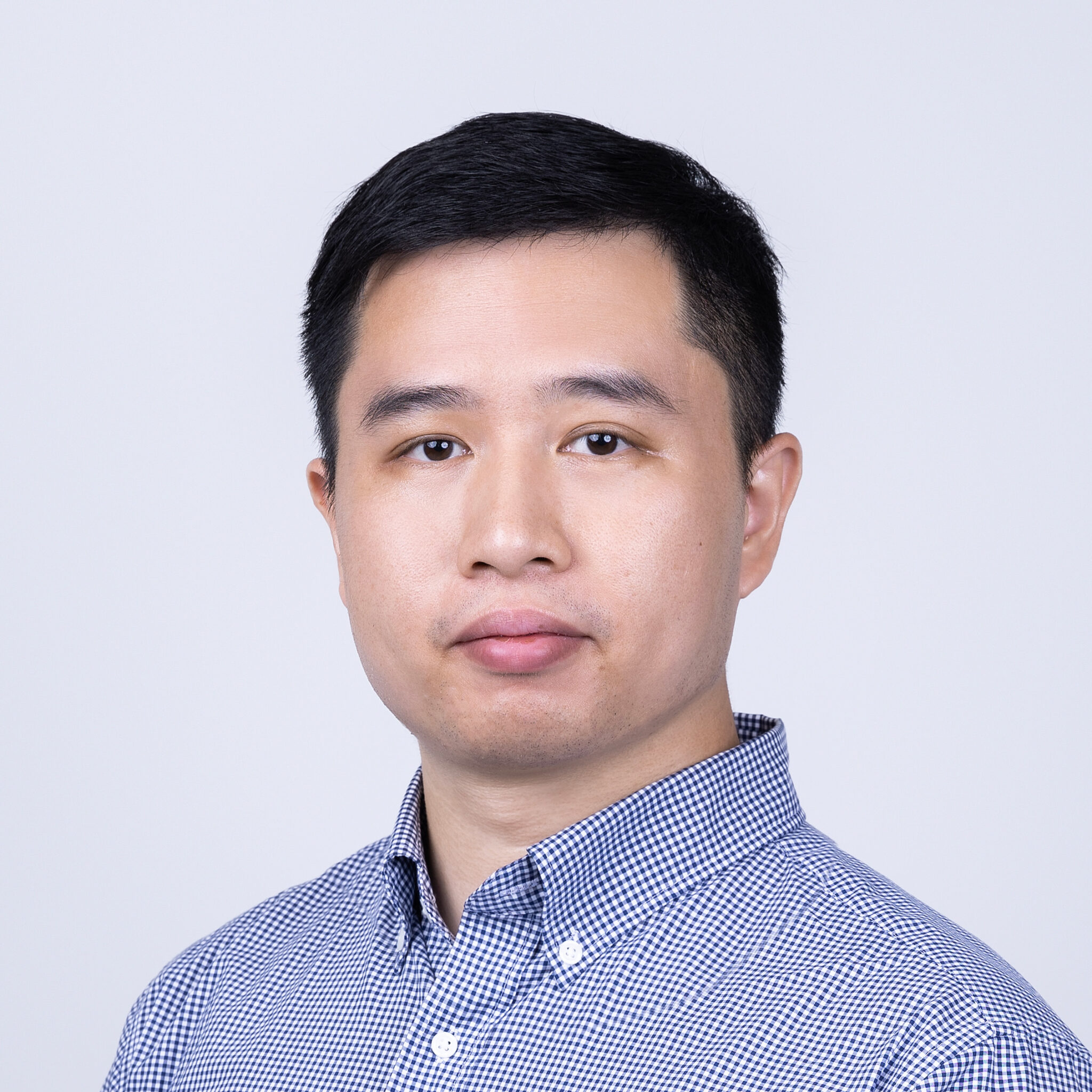}}]{Difan Zou} is an Assistant Professor in HKU IDS \& Computer Science, School of Computing and Data Science, at The University of Hong Kong. He received his Ph.D. in Computer Science, University of California, Los Angeles (UCLA). He received a B. S degree in Applied Physics, from School of Gifted Young, USTC and a M. S degree in Electrical Engineering from USTC. He has published multiple papers on top-tier machine learning conferences including ICML, NeurIPS, ICLR, COLT, etc. He is a recipient of Bloomberg Data Science Ph.D. fellowship. His research interests are broadly in machine learning, optimization, and learning structured data (e.g., time-series or graph data), with a focus on theoretical understanding of the optimization and generalization in deep learning problems.
\end{IEEEbiography}

\begin{IEEEbiography}[{\includegraphics[width=1in,height=1.25in,clip]{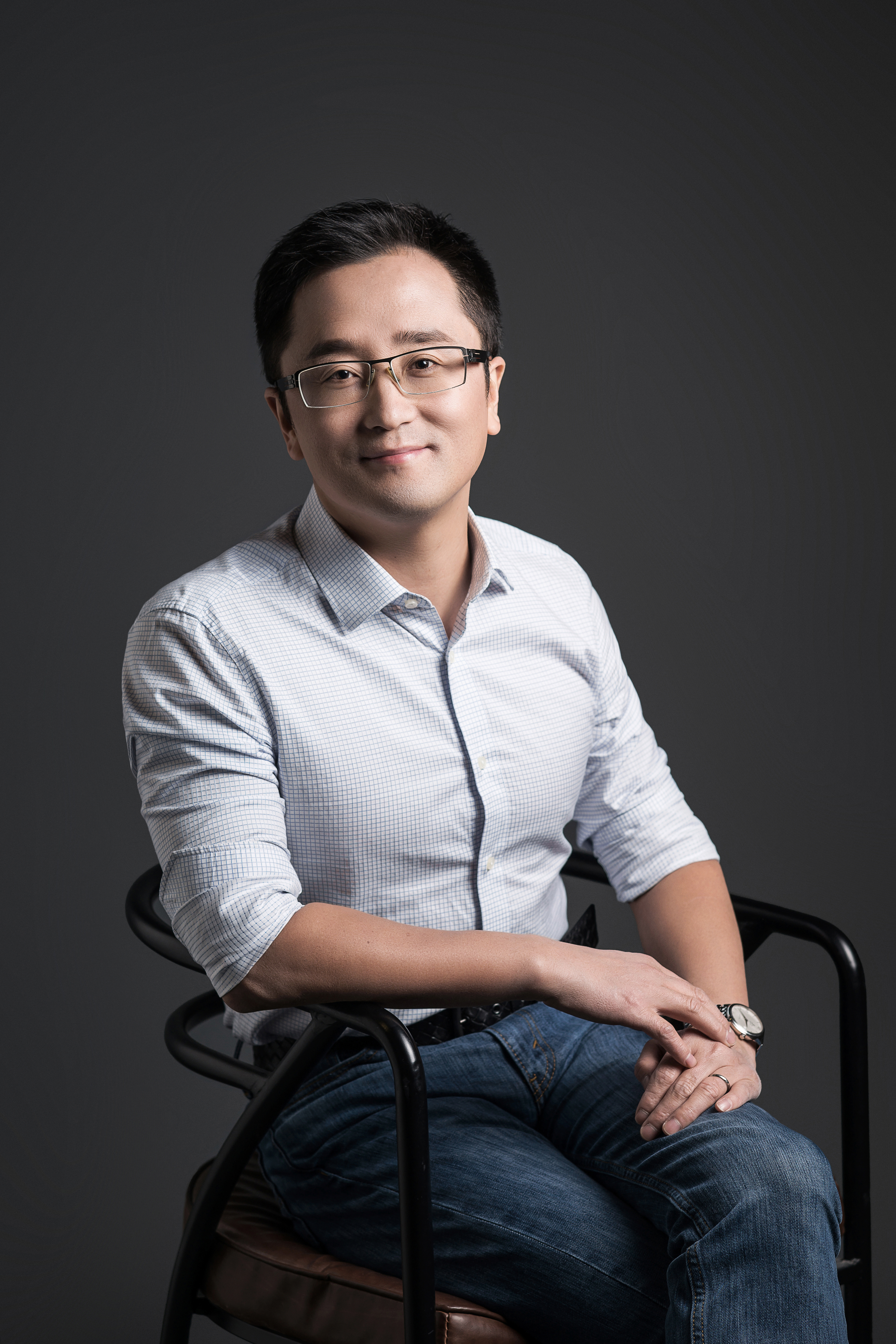}}]{Heng Tao Shen} is a professor with the School of Computer Science and Technology, Tongji University, China. 
He obtained his BSc with 1st class Honours and PhD from Department of Computer Science, National University  of  Singapore in 2000 and 2004 respectively.
His current research interests include multimedia search, computer vision, artificial intelligence, and big data management. He has published 300+ peer-reviewed papers and received 7 best paper awards from international conferences, including the Best Paper Award from ACM Multimedia 2017 and Best Paper Award-Honourable Mention from ACM SIGIR 2017. He has served as General Co-chair for ACM Multimedia 2021 and TPC Co-Chair for ACM Multimedia 2015, and is an Associate Editor of ACM Trans. of Data Science (TDS), IEEE Trans. on Image Processing (TIP), IEEE Trans. on Multimedia (TMM), and IEEE Trans. on Knowledge and Data Engineering (TKDE). He is a Fellow of ACM/IEEE/OSA.
\end{IEEEbiography}

\begin{IEEEbiography}[{\includegraphics[width=1in,height=1.25in,clip]{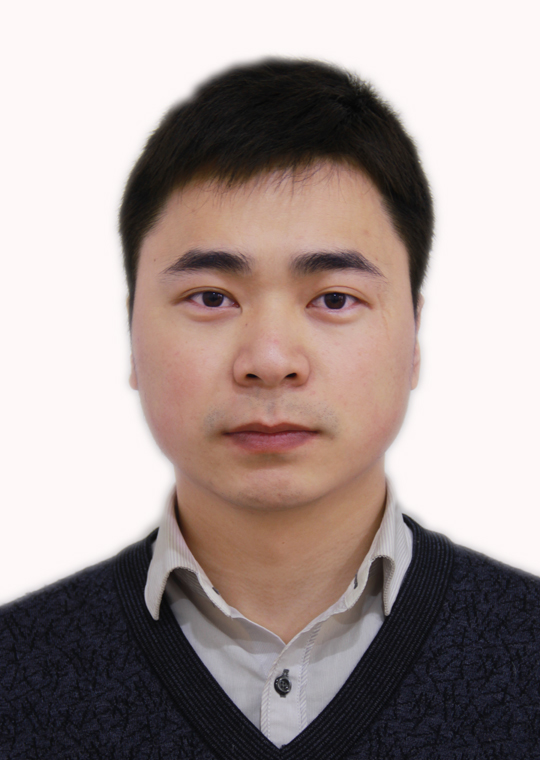}}]{Jingkuan Song} is a professor with the School of Computer
Science and Technology, Tongji University, China. 
He joined Columbia University as a Postdoc Research Scientist (2016-2017), and University of Trento as a Research Fellow (2014-2016). He obtained his PhD degree in 2014 from The University of Queensland (UQ), Australia. His research interest includes large-scale multimedia retrieval, LLMs and deep learning techniques. He was the winner of the Best Paper Award in ICPR (2016, Mexico), Best Student Paper Award in Australian Database Conference (2017, Australia), and Best Paper Honorable Mention Award (2017, Japan). He is an Associate Editor of IEEE TMM and ACM TOMM.
\end{IEEEbiography}

\end{document}